%% file: main.tex
\documentclass[runningheads]{llncs}

\usepackage[final]{eccv}

\usepackage{eccvabbrv}

\usepackage{graphicx}
\usepackage{booktabs}
\usepackage{tabularx}
\usepackage[ruled,vlined]{algorithm2e}

\def\SAEcolor{\textcolor[RGB]{154,25,254}}
\definecolor{mygray}{gray}{.9}
\definecolor{commentcolor}{RGB}{110,154,155}
\usepackage{xcolor}
\usepackage{colortbl}
\usepackage{marvosym}

\usepackage{multirow}
\usepackage[accsupp]{axessibility} 

\usepackage[pagebackref,breaklinks,colorlinks,citecolor=eccvblue]{hyperref}

\usepackage{orcidlink}

\begin{document}

\title{Shifted Autoencoders for Point Annotation Restoration in Object Counting} 


\author{Yuda Zou\textsuperscript{*}, ~Xin Xiao\textsuperscript{*}, ~Peilin Zhou, ~Zhichao Sun, ~Bo Du, ~Yongchao Xu\textsuperscript{(\Letter)}}

\authorrunning{Y. Zou et al.}

\institute{School of Computer Science, Wuhan University, Wuhan, China\\
\email{\{zouyuda,xinxiao,peilinzhou,zhichaosun,dubo,yongchao.xu\}@whu.edu.cn}}

\maketitle

\begingroup
\renewcommand\thefootnote{}\footnote{\textsuperscript{*}equal contribution}
\addtocounter{footnote}{-1}
\endgroup

\input{paper_sections/0_abstract}    
\input{paper_sections/1_introduction}

\input{paper_sections/2_related_work}

\input{paper_sections/3_methodology}

\input{paper_sections/4_experimentation}
\input{paper_sections/5_conclusion}

\bibliographystyle{splncs04}
\bibliography{reference}

\input{paper_sections/X_supple}

\end{document}

%% file: paper_sections/0_abstract.tex
\begin{abstract}

Object counting typically uses 2D point annotations. 
The complexity of object shapes and the subjectivity of annotators may lead to annotation inconsistency, potentially confusing counting model training. Some sophisticated noise-resistance counting methods have been proposed to alleviate this issue. Differently, we aim to directly refine the initial point annotations before training counting models. For that, we propose the Shifted Autoencoders (SAE), which enhances annotation consistency.
Specifically, SAE applies random shifts to initial point annotations and employs a UNet to restore them to their original positions. 
Similar to MAE reconstruction, the trained SAE captures general position knowledge and ignores specific manual offset noise. This allows to restore the initial point annotations to more general and thus consistent positions. Extensive experiments show that using such refined consistent annotations to train some advanced (including noise-resistance) object counting models steadily/significantly boosts their performances.
Remarkably, the proposed SAE helps to set new records on nine datasets. We will make codes and refined point annotations available. 
  \keywords{Object counting \and Annotation refinement \and Crowd counting}
  
\end{abstract}

%% file: paper_sections/1_introduction.tex
\section{Introduction}
\label{sec: introduction}

Object counting~\cite{li2018csrnet, liu2019context_can, xu2019learn_autoscale_iccv, song2021rethinking_P2P, ma2019bayesian_BL, liang2023crowdclip, gao2020counting_ASPD_RSOC, chen_direction_field}, increasingly vital in domains like security surveillance~\cite{sindagi2017cnn_surveillance}, urban planning~\cite{urban_planning}, and biological research~\cite{paper_biomedical}, has benefited greatly from advancements in computer vision. Most object counting methods can be roughly classified into two categories: localization-based~\cite{sam2020locate_box, liu2023point_PET, song2021rethinking_P2P} and density-map-based approaches~\cite{ma2019bayesian_BL, wan2020modeling_NoiseCC, huang2023counting_AWCC, rsi_cheng2022rethinking, lin2022boosting_MAN, gao2020counting_ASPD_RSOC, gao2022psgcnet, ding2022object_ADMAL, guo2021sau, wan2021fine, sun2023indiscernible_underwater}. Localization-based methods focus on identifying individual objects with bounding box~\cite{sam2020locate_box} or point~\cite{song2021rethinking_P2P, liu2023point_PET, xu2022autoscale_ijcv} representation. In contrast, density-map-based methods apply regression techniques to estimate the density distribution of objects. 

Object counting datasets~\cite{zhang2016single_SH_MCNN, idrees2018composition_UCF-QNRF, sindagi2020jhu, gao2020counting_ASPD_RSOC, paper_MBM, paper_ADI, paper_DCC}, distinct from those for object detection, predominantly use 2D coordinate points for marking objects. This annotation mode is particularly advantageous for densely packed or overlapping objects. However, this approach inevitably results in variations and inconsistencies in point annotations within the dataset, primarily due to the subjective decisions made by annotators in selecting the point annotation positions for different objects of the same type. Inconsistencies in point annotations may introduce ambiguity and confusion during the training phase of counting models, compromising their counting accuracy.

\begin{figure}[t]
  \centering
  \includegraphics[width=0.92\linewidth]{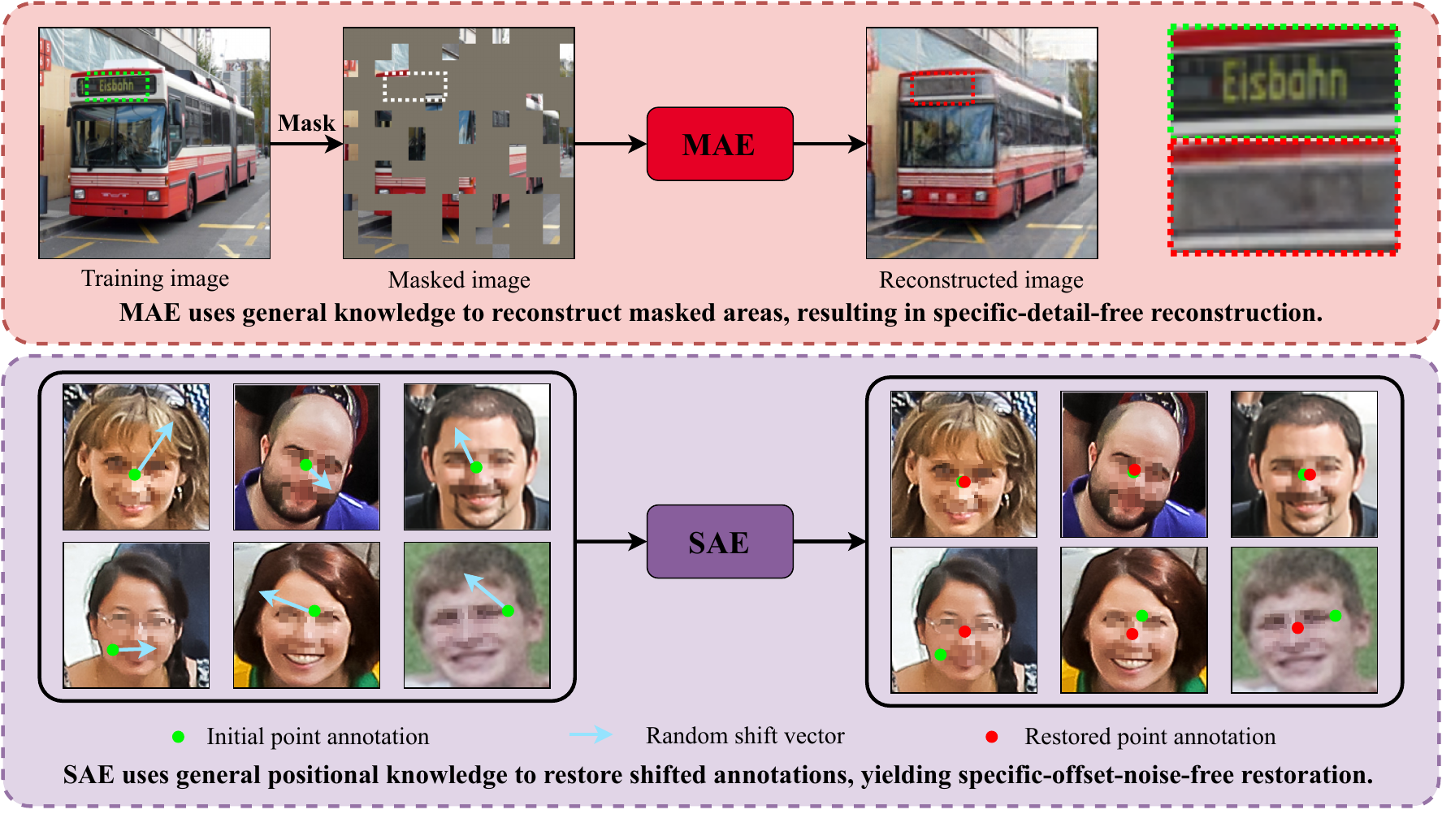}
  \caption{Drawing inspiration from MAE, our SAE captures general positional knowledge by being trained to restore the shifted point annotations to their original positions. In the restoration phase, the trained SAE restores the initial point annotations to more common positions using the learned general positional knowledge.}
  \label{figure: motivation}
\end{figure}

To mitigate the issue of annotation inconsistencies in object counting datasets, various strategies have been proposed. ADSCNet~\cite{Bai_2020_CVPR_ADSC} employs prediction of the network to adjust the target density map using the Gaussian Mixture Model (GMM)~\cite{reynolds2009gmm}. Other methods like BL~\cite{ma2019bayesian_BL} and RSI~\cite{rsi_cheng2022rethinking} focus on enhancing noise resistance by altering the loss function and convolution filter, respectively. Building upon BL, NoiseCC~\cite{wan2020modeling_NoiseCC} introduces a novel loss function that lowers the weights of uncertain regions on the density map, thereby reducing the influence of noisy annotations. 
These methods, though effective in enhancing the model's tolerance to annotation inconsistency, are primarily involved in the training phase based on the inconsistent annotations. Integration of these methods into other frameworks may complicate the training process. Additionally, their application to localization-based methods remains unverified, as they were originally designed for density-map-based approaches. 
Alternatively, directly improving initial annotations presents a more efficient and broadly applicable solution.

In this paper, we aim to directly refine the inconsistent point annotations to consistent ones. Drawing inspiration from Masked Autoencoders (MAE)~\cite{he2022mae},
we introduce the Shifted Autoencoders (SAE). Specifically,
as illustrated in Fig.~\ref{figure: motivation}, MAE is trained by masking a portion of an image and reconstructing the masked image to the original unmasked one. As a form of Denoised Autoencoders~\cite{DAE}, the MAE captures the general knowledge and discards the specific knowledge of all the training data through the reconstruction training process on a vast number of images~\cite{DAE}. As a result, the trained MAE tends to reconstruct generic patterns and ignore specific patterns even for previously trained images (see the reasonable overall representation of the reconstructed bus and the missing text on the bus of Fig.~\ref{figure: motivation}). From the aspect of reconstruction quality, the image reconstructed by MAE is not perfectly the same as the original one. \textbf{Interestingly, such a similar imperfect restoration is what we need for point annotation restoration.}

Similar to MAE~\cite{he2022mae} that applies random masks to the original image and aims to reconstruct it, we apply random 2D shifts to the initial point annotations and predict a restoration vector field to restore these shifted point annotations to their original positions. While trained on numerous similar objects and their point annotations, the SAE, akin to MAE, is compelled to capture the general positional knowledge and discard the specific manual offset noise knowledge of all the manual annotations in the training data.
In the restoration phase, the trained SAE regards the initial manual point annotations as shifted points and uses learned general positional knowledge to restore them. In this way, the restored point annotations mitigate their individual manual offset noise, presenting more consistency with each other (see the bottom of Fig.~\ref{figure: motivation}).

We conduct extensive experiments on eleven datasets of three applications (crowd counting, remote sensing object counting, and cell counting). 
Compared to training with the initial point annotations, training with the ones revised by SAE steadily boosts the performance of some state-of-the-art counting methods.

The main contributions of this work are threefold: 1) We present the idea of directly refining the point annotations to be more consistent, which is beneficial for both density-map-based and localization-based object counting methods. 2) We novelly introduce the Shifted AutoEncoders (SAE), which effectively captures the general positional knowledge and ignores specific manual offset noise within the training data, yielding consistent point annotations. 3) The proposed SAE helps to set new state-of-the-art results on nine datasets.

%% file: paper_sections/2_related_work.tex
\section{Related work}
\label{sec: related work}

There are roughly two types of object counting methods: density-map-based methods~\cite{wan2020modeling_NoiseCC, huang2023counting_AWCC, rsi_cheng2022rethinking, gao2020counting_ASPD_RSOC, gao2022psgcnet, ding2022object_ADMAL, guo2021sau} 
and localization-based methods~\cite{sam2020locate_box, liu2023point_PET, song2021rethinking_P2P}.
Both types of methods heavily rely on the point annotation quality. Existing methods~\cite{Bai_2020_CVPR_ADSC, ma2019bayesian_BL, wan2020modeling_NoiseCC, rsi_cheng2022rethinking, wan2023modeling_NoiseCC_Tpami} coping with the inconsistent point annotations mainly focus on enhancing the model's tolerance to annotation noise. 
We shortly review these methods in the following.

\begin{figure}[t]
  \centering
  \includegraphics[width=0.8\linewidth]{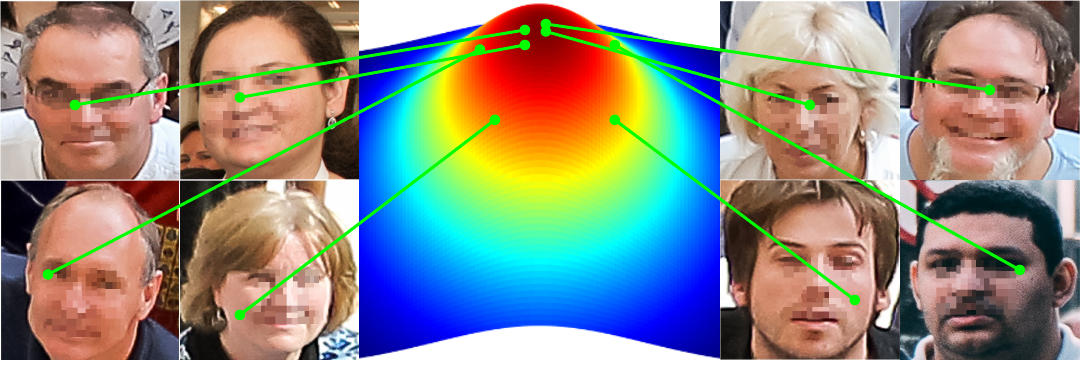}
  \caption{Illustrative example of relative spatial distribution (approximating a Gaussian distribution~\cite{wan2023modeling_NoiseCC_Tpami}) of point annotations \textit{w.r.t.} corresponding heads.}
  \label{figure: gaussian}
\end{figure}

\subsection{Density-map-based object counting}
\label{subsec: related_work_density_based_object_counting}
Density-map-based methods have emerged as the predominant approach in object counting. These methods involve creating a learning target in the form of a density map, which is constructed by applying Gaussian kernels to smooth point annotations.  Models are then trained to emulate this map, with counting results derived through spatial integration over the density map. Recent improvements in this field include the creation of more complex network architectures~\cite{lin2022boosting_MAN, li2018csrnet, gao2022psgcnet, xu2019learn_autoscale_iccv}, refinement of loss functions~\cite{ma2019bayesian_BL, wan2020modeling_NoiseCC, wan2021generalized_GL, idrees2018composition_UCF-QNRF}, introduction of new density map formats~\cite{liang2022focal_FIDTM, wan2020kernel_based, ding2022object_ADMAL, wan2019adaptive_density}, and the integration of scale variability factors~\cite{zhang2016single_SH_MCNN, song2021choosefuse, gao2020counting_ASPD_RSOC}. While density-map-based methods have achieved significant success, they are limited in providing individual object location information.

\subsection{Localization-based object counting}
\label{subsec: related_work_localization_based_object_counting}
Localization-based object counting methods~\cite{sam2020locate_box, liu2023point_PET, song2021rethinking_P2P} focus on directly pinpointing each target object, offering broader applicability. Early methods regard counting as an object detection problem using pseudo bounding boxes~\cite{sam2020locate_box}, and struggle in congested areas. Recent advancements like P2PNet~\cite{song2021rethinking_P2P} and PET~\cite{liu2023point_PET} go beyond bounding boxes. P2PNet employs point localization through Hungarian matching~\cite{kuhn1955hungarian} with fixed anchor points. In contrast, PET strategically places anchor points, further improving the counting accuracy. While localization-based methods offer detailed object localization information, they may not perform well in extremely dense areas.

\subsection{Methods focusing on annotation inconsistencies}
\label{subsec: related_work_annotation_inconsistencies}

ADSCNet~\cite{Bai_2020_CVPR_ADSC} introduces a novel framework that leverages network prediction to refine target density map using Gaussian Mixture Models (GMM). Both BL~\cite{ma2019bayesian_BL} and RSI~\cite{rsi_cheng2022rethinking} emphasize enhancing noise resistance in their respective frameworks. In particular, BL achieves this through the introduction of an innovative loss function. RSI redesigns the convolutional filter. Building upon BL, NoiseCC~\cite{wan2020modeling_NoiseCC} incorporates a loss function that reduces the loss weights of uncertain regions on the density map, mitigating the impact of noisy annotations.

These methods primarily focus on enhancing the model's tolerance to annotation noise rather than directly improving the quality of the initial point annotations. Additionally, applying these methods each time incurs additional training costs. Furthermore, these methods are primarily tailored for density-based methods, which may impose limitations on their applicability to localization-based methods. In contrast, our proposed approach focuses on directly refining initial point annotations, presenting a more efficient and broadly applicable solution.

\begin{figure}[t]
  \centering
  \includegraphics[width=1.0\linewidth]{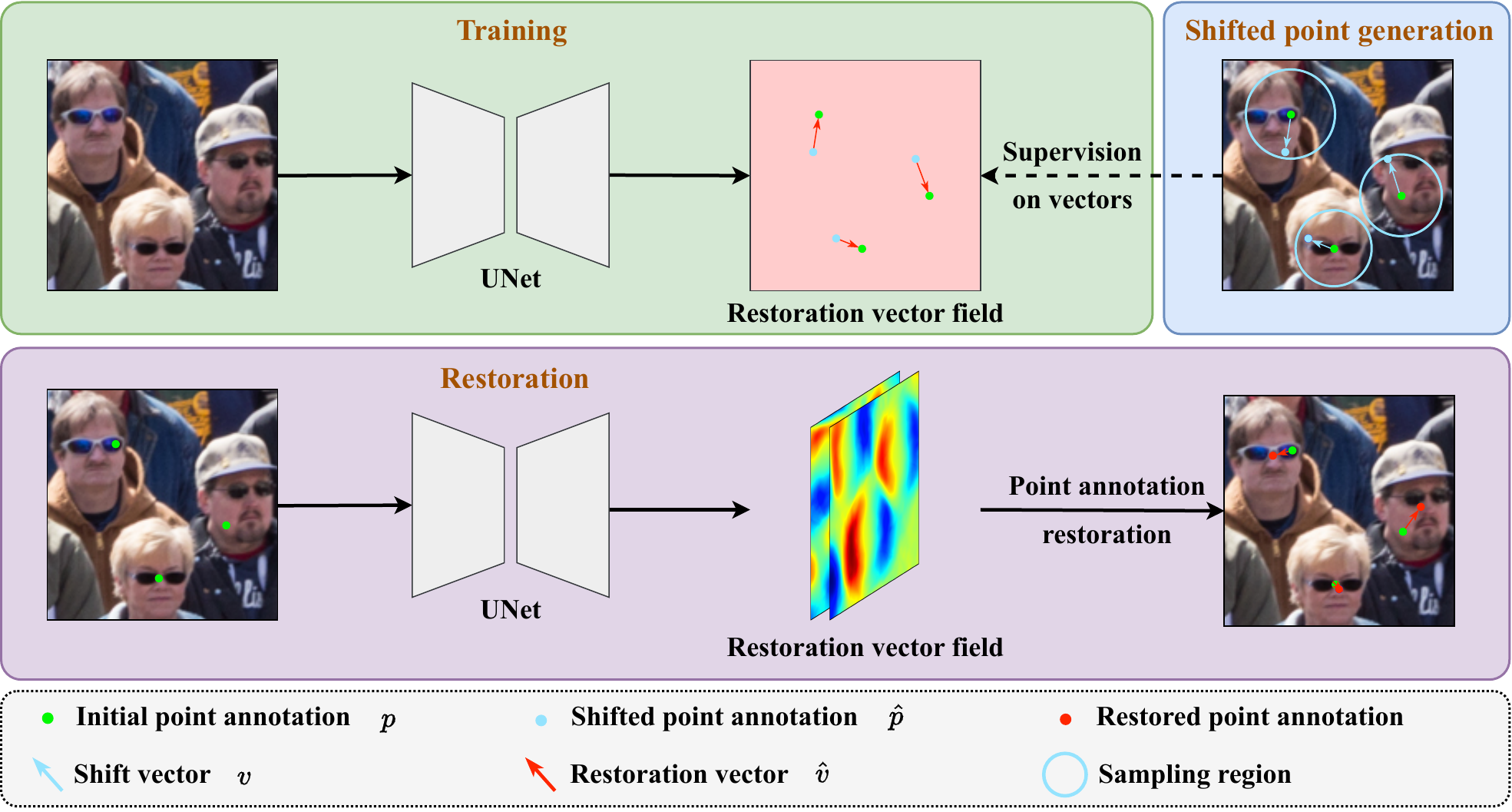}
  \caption{The pipeline of the proposed Shifted Autoencoders (SAE), consisting of three steps: 1) Shifted point generation by adding random shift vectors to the initial annotated points; 2) Training the SAE with generated shift vectors by restoring shifted points to their original positions based on predicted restoration vector field; 
  3) Self-restoration that shifts the originally annotated points with corresponding restoration vectors in the predicted vector field.}
  \label{figure: pipeline}
\end{figure}

%% file: paper_sections/3_methodology.tex
\section{Methodology}
\label{sec: Methodology}

\subsection{Overview}
\label{subsec: methodology_overview}
The quality of point annotation affects the training of counting models, which is vital for the final counting accuracy. Yet, the subjective decisions made by annotators in selecting the point annotation positions often lead to inconsistent annotation. Taking crowd counting as an example, the relative spatial distribution of manual point annotations \textit{w.r.t.} corresponding objects (heads) roughly obeys a Gaussian function~\cite{wan2023modeling_NoiseCC_Tpami}. As shown in Fig.~\ref{figure: gaussian}, the manual point annotations share a common distribution center (representing general position), and have specific and different offsets from this center (indicating specific manual offset noise). Inspired by Masked Autoencoders (MAE)~\cite{he2022mae}, we propose the Shifted Autoencoders (SAE), a novel approach designed to improve the consistency of point annotations within object counting datasets before training counting models.

The workflow of SAE comprises three steps (see Fig.~\ref{figure: pipeline}).
Firstly, we generate shifted point annotations by applying random shifts to the initial point annotations. 
Secondly, the SAE network is trained to restore these shifted point annotations to their original positions by predicting a restoration vector field. While trained on numerous similar objects and their point annotations, the SAE, similar to MAE~\cite{he2022mae}, is compelled to capture the general positional knowledge and ignores the specific manual offset noise knowledge of all the manual annotations in training data. Finally, we regard initial manual point annotations as shifted point annotations, and adopt the trained SAE to restore them toward the common distribution center, yielding more consistent annotations.

\subsection{Shifted point annotation generation}
\label{subsec: methodology_generation}

Given an image $X$ in the training set, it is associated with a set of point annotations $P = \{{p_i = (x_i, y_i)}\}$, where $x_i$ (\textit{resp.} $y_i$) represents the $x$ (\textit{resp.} $y$) coordinate of the $i$-th point annotation $p_i$. The $i$ ranges from 1 to $N$, where $N$ denotes the total number of point annotations within this image. For each annotation point $p_i = (x_i, y_i)$ in $P$, we apply a random 2D shift vector $v_i$, which can be decomposed into two separate components: angle $a_i$ and magnitude $m_i$. Each component is considered independently. Specifically, we uniformly sample from $(0, 2\pi)$ for the angle $a_i$.

To ensure that shift vectors remain within a reasonable range like the limited masking ratio of MAE~\cite{he2022mae}, we set an upper bound $r_i$ for the magnitude $m_i$. For simplicity, we uniformly sample from $(0, r_i)$ for $m_i$. In the following, we detail how to set $r_i$ and generate shifted point annotation.

\medskip
\noindent
\textbf{Radius of sampling region for shift vector.}
As depicted in Fig.~\ref{figure: pipeline}, the shift vector $v_i$ is the vector pointing from the initial point annotation $p_i$ towards the shifted point annotation $\hat{p_i}$. This means the upper bound, $r_i$, for the magnitude of $v_i$ is actually the radius of the circular region within which the $\hat{p_i}$ is randomly sampled.
Naturally, the radius $r_i$ is expected to equal the scope radius of the corresponding object in the ideal case.

\begin{figure}[t]
\centering
\subfloat[Sampling regions with $r = \alpha \times d$]{\includegraphics[width=.49\linewidth]{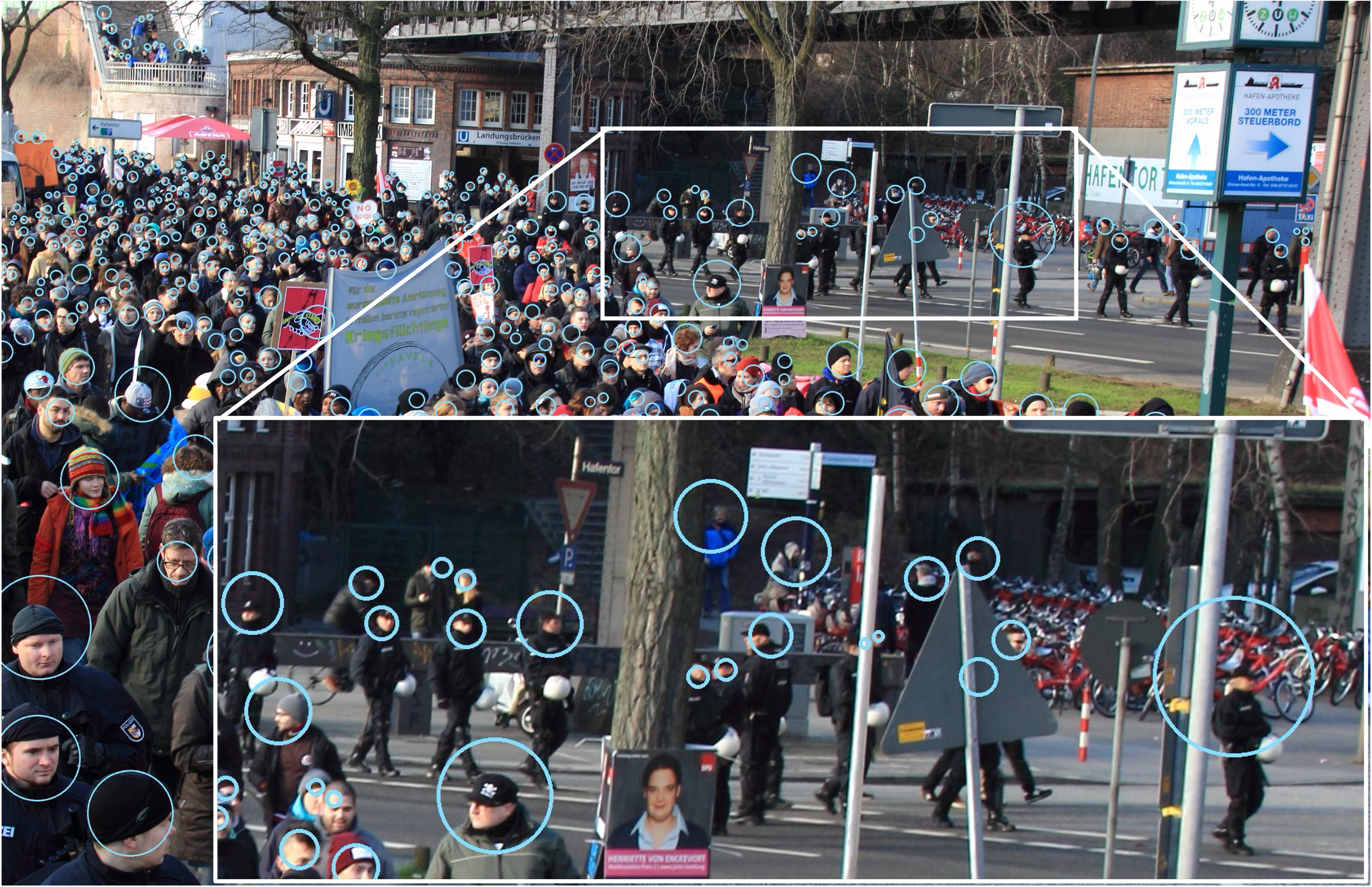}\label{figure: sample_region_results_with_only_distance}}\hspace{0.1pt}
\subfloat[Sampling regions with $r = \alpha \times min(d, \overline{d}^{\mathcal{N}_3})$]{\includegraphics[width=.49\linewidth]{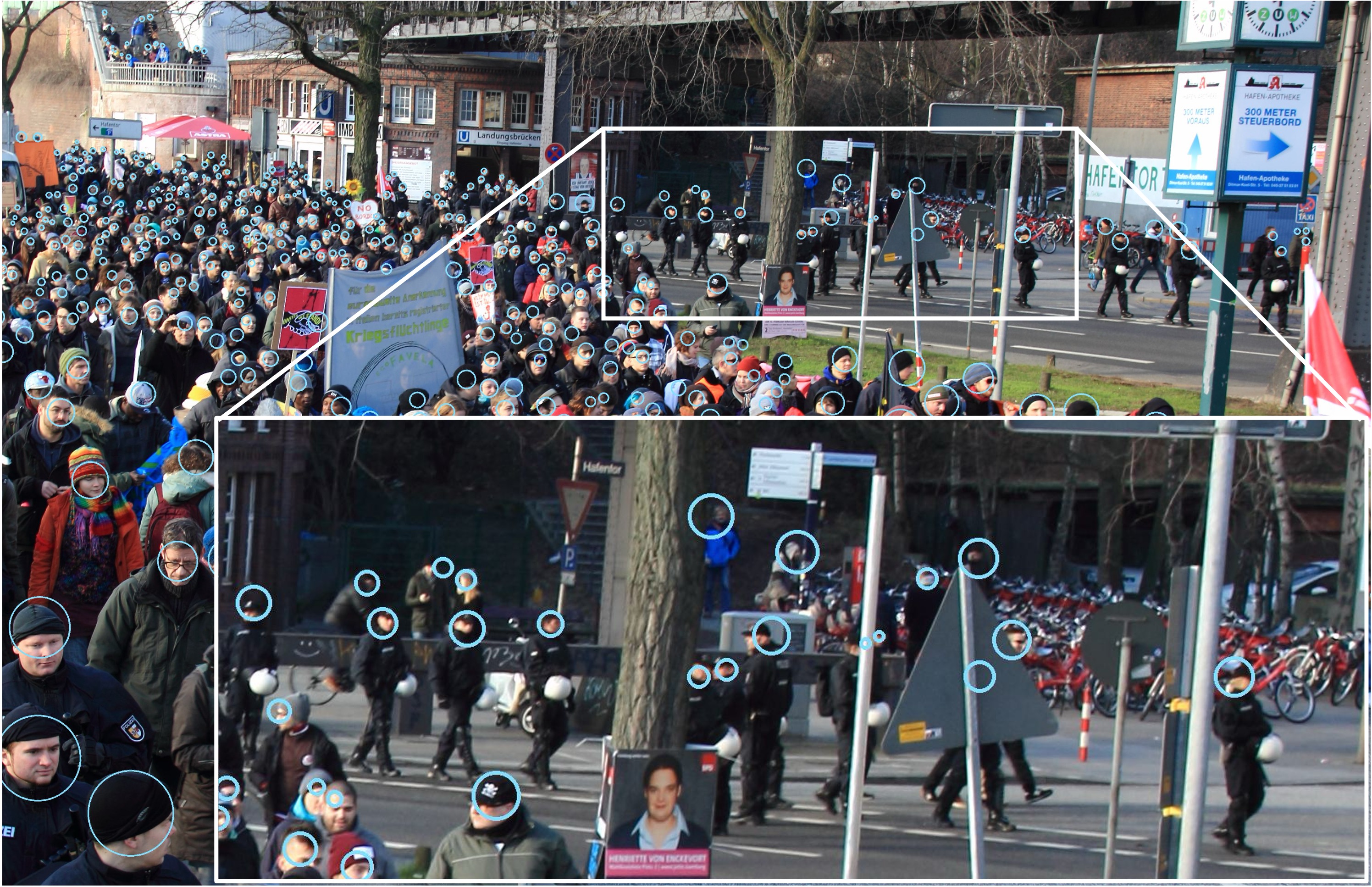}\label{figure: sample_region_results_with_distance_and_limitation}}

\caption{Illustration of different strategies to define the radius of sampling region for each annotated point in crowd images. See Eq.~\eqref{eq: radius_with_only_distance} and Eq.~\eqref{eq: radius_with_distance_and_limitation} and corresponding text for more details. The hyper-parameter $\alpha$ is set to 0.4. Best viewed by zooming in the electronic version. }
\label{figure: sample_region_results}
\end{figure}

Object counting annotations are typically in the form of two-dimensional points, which provide no direct information about the sizes or shapes of the corresponding objects.
These 2D point annotations can only provide information about the spatial distribution of these objects and the distances between them on the image plane. 
For object counting, some density-map-based methods~\cite{zhang2016single_SH_MCNN, li2018csrnet} approximate the size of objects of the same type based on the distances between them. 
Following them, a straightforward solution to roughly determine the size of objects of the same type is to leverage the distance information between them.  
Specifically, for each point annotation $p_i$ in $P$, we denote the Euclidean distance to its nearest neighbor as $d_i$. 
As stated in~\cite{zhang2016single_SH_MCNN}, the pixel associated with the $i$-th object corresponds to an area roughly of a radius proportional to $d_i$ 
Therefore, a simple choice to set the radius $r_i^s$ is given by:
\begin{equation}
\label{eq: radius_with_only_distance}
r_i^s \, = \, \alpha \times d_i,
\end{equation}
where $\alpha$ is a hyper-parameter which is constrained not to exceed 0.5. As illustrated in Fig.~\ref{figure: sample_region_results_with_only_distance}, this 0.5 constraint prevents the overlapping of sampling regions with neighboring point annotations, which may bring confusion on the target point $p_i$, towards which the shifted point $\hat{p}_i$ should be restored.

This simple setting for the radius $r_i$ is acceptable to approximate the size for most objects (see Fig.~\ref{figure: sample_region_results_with_only_distance}). However, for some objects that are distributed sparsely, the radius obtained based on Eq.~\eqref{eq: radius_with_only_distance} would be significantly oversized. This is due to their distances to the corresponding nearest neighbors being much greater than their spatial size. 
As illustrated in the white rectangle of Fig.~\ref{figure: sample_region_results_with_only_distance}, an oversized sampling radius will introduce an excessive amount of background to the sampling region.
For these objects, the proposed SAE will be forced to consider the background areas far from the objects as a part of them, which compromises the training of SAE.

To mitigate this, we comprehensively take into account the size information of neighboring objects of $p_i$.
Specifically, the final radius $r_i$ for the annotation point $p_i$ is defined by: 
\begin{equation}
\label{eq: radius_with_distance_and_limitation}
r_i = \alpha \times min(d_i, \overline{d}_i^{\mathcal{N}_3}),
\end{equation}
where $\overline{d_i}^{\mathcal{N}_3}$ denotes the mean value of $d$ for the three nearest neighboring annotation points of $p_i$. 
By combining $d$ of neighboring point annotations, the sampling regions of the sparsely distributed point annotations are better aligned with the spatial scales of their objects. An illustration example is shown in the white rectangle of Fig.~\ref{figure: sample_region_results_with_distance_and_limitation}.


\medskip
\noindent\textbf{Shift vector and shifted point annotation.}
With the angle $a_i$ and the magnitude $m_i$, the 2D shift vector $v_i$ for the annotation point $p_i$ can be generated as below:
\begin{equation}
    v_i = \begin{pmatrix}
    v_i^x, v_i^y\\
    \end{pmatrix} = \begin{pmatrix}
    m_i \times \cos{a_i}, m_i \times \sin{a_i}\\
    \end{pmatrix}
\label{eq: shift_vector}
\end{equation}
By imposing $v_i$ to the corresponding initial point annotation $p_i$, the shifted point annotation $\hat{p_i}$ is obtained as follows: 
\begin{equation}
    \hat{p_i} =   (\hat{x_i}, \hat{y_i})  =(x_i + v_i^x, y_i + v_i^y)
\label{eq: shifted_point annotation}
\end{equation}

\subsection{SAE network training}
\label{subsec: methodology_training}
\textbf{SAE network architecture.} In our SAE network, we employ a lightweight UNet architecture~\cite{ronneberger2015unet}, incorporating a VGG16~\cite{simonyan2014very_vgg} backbone, to predict the restoration vector field $F \in \mathbb{R}^{2 \times H \times W}$.
This field consists of two channels, $F_x \in \mathbb{R}^{H \times W}$ and $F_y \in \mathbb{R}^{H \times W}$, corresponding to the $x$ and $y$ axes, respectively. Here, $H$ and $W$ denote the height and the width of the input image, respectively.

\smallskip
\noindent\textbf{Training objective.} In our SAE, we aim to restore the shifted point annotations $\{\hat{p}_i\}$ generated in Sec.~\ref{subsec: methodology_generation} to their corresponding initial positions $\{p_i\}$. In alignment with the approach adopted in MAE~\cite{he2022mae}, we adopt the Mean Squared Error (MSE) loss to train our SAE network. Note that we only compute the MSE loss with $F$ on the coordinates of shifted point annotations $\{\hat{p}_i\}$. Specifically, the restoration shift vectors $\{\hat{v}_i \, = \, \big(F_x(\hat{p}_i), F_y(\hat{p}_i)\big)\}$ are supervised to align with the inverse of the corresponding shift vectors $\{v_i\}$. Formally, the training objective $\mathcal{L}$ for our SAE is given by:
\begin{equation}
\label{eq: loss}
\mathcal{L} \, = \, \frac{1}{N} \sum_{i=1}^N\left\|\hat{v}_i-(- v_i)\right\|^2,
\end{equation}
where $N$ refers to the total number of annotated points in a given image within the training set. 

After training on a variety of similar objects and their corresponding point annotations, the SAE, similar to MAE~\cite{he2022mae}, captures the general positional knowledge while discarding the specific manual offset noise of all the manual point annotations in the training data. 
Indeed, imaging a synthetic case where the same head repeats multiple times with different point annotations spreading around a fixed position (e.g., head center), the optimal choice to simultaneously cope with the varied annotations is to restore them to the same fixed position. In practice, optimizing the SAE network on a large number of objects also tends to restore the manual annotations towards the distribution center, yielding more consistent annotation.

\subsection{Point annotation restoration}
\label{subsec: methodology_restoration}
The trained SAE network is then utilized to revise the initial point annotations. More precisely,
each image from the training set is processed through the trained SAE network to generate the corresponding 2D restoration vector field map $F$. 
The initial point annotations are treated as shifted point annotations and are restored by the predicted restoration vector field map in a manner analogous to the training phase.
Specifically, 
the coordinate of the initial point annotation, $p_i$, is used to sample the corresponding restoration vector $\hat{v_i}$ within $F$. The final restored point annotations are obtained by adding the restoration vectors to the corresponding initial point annotations. Formally, the set of restored point annotations $P_r$ is given by:
\begin{equation}
\label{eq: restored_or_revised_point_annotation}
    P_{r} =\{(x_i + \hat{v}_i^x, y_i + \hat{v}_i^y)\} = \{\big(x_i + F_x(p_i), y_i + F_y(p_i)\big)\}
\end{equation}

Thanks to the captured general positional knowledge and avoidance of specific manual offset noise of our SAE, the restored point annotations have better consistency among themselves. 
Compared with training with the initial point annotations, using more consistent point annotations introduces less confusion during the training of counting models, resulting in improved counting accuracy.

%% file: paper_sections/4_experimentation.tex
\section{Experiments}
\label{sec: experimentation}

\subsection{Experimental setting}
\label{subsec: experimental_setting}
\noindent\textbf{Datasets.} To evaluate the effectiveness of the proposed approach, we perform extensive experiments on eleven datasets spanning three domains: crowd counting, remote sensing object counting, and cell counting. These datasets include: SH PartA~\cite{zhang2016single_SH_MCNN}, SH PartB~\cite{zhang2016single_SH_MCNN}, UCF-QNRF~\cite{idrees2018composition_UCF-QNRF}, JHU++~\cite{sindagi2020jhu}, RSOC\_building~\cite{gao2020counting_ASPD_RSOC}, RSOC\_small-vehicle~\cite{gao2020counting_ASPD_RSOC}, RSOC\_large-vehicle~\cite{gao2020counting_ASPD_RSOC}, RSOC\_ship~\cite{gao2020counting_ASPD_RSOC}, MBM~\cite{paper_MBM}, ADI~\cite{paper_ADI}, and DCC~\cite{paper_DCC}.

\smallskip
\noindent\textbf{Evaluation metrics.} 
In object counting tasks, the Mean Absolute Error (MAE) and Mean Square Error (MSE) serve as the principal evaluation metrics. Lower values for these metrics indicate better counting performance.

\smallskip
\noindent\textbf{Implementation details.}  
During the training phase for SAE, we augment the images with random scaling, horizontal flipping, and randomly cropping to $512 \times 512$ pixels (except for the ADI dataset~\cite{paper_ADI} which uses $128 \times 128$ pixels due to limited resolution). Shift vectors for point annotations are randomly regenerated at each iteration.
We employ the Adam optimizer with a weight decay of $5 \times 10^{-4}$ and a fixed learning rate of $1 \times 10^{-4}$. SAE training proceeds for 100 epochs with batches of 8 for each dataset on an RTX 3090 GPU using the PyTorch framework. To ensure a fair comparison, we train the SAE on the training set of each dataset individually, thereby avoiding the introduction of extra data. It is noteworthy that the proposed \textbf{SAE plays the role only in the point annotation restoration phase, rather than the object counting phase}. During the training phase for object counting, we follow the same implementation details as the baseline counting methods. The notation \& \SAEcolor{SAE} below indicates that the counting methods are trained using point annotations revised by SAE. Otherwise, they are trained with the initial manual point annotations.

\begin{figure}[t]
\centering
\subfloat{\includegraphics[width=.24\linewidth]{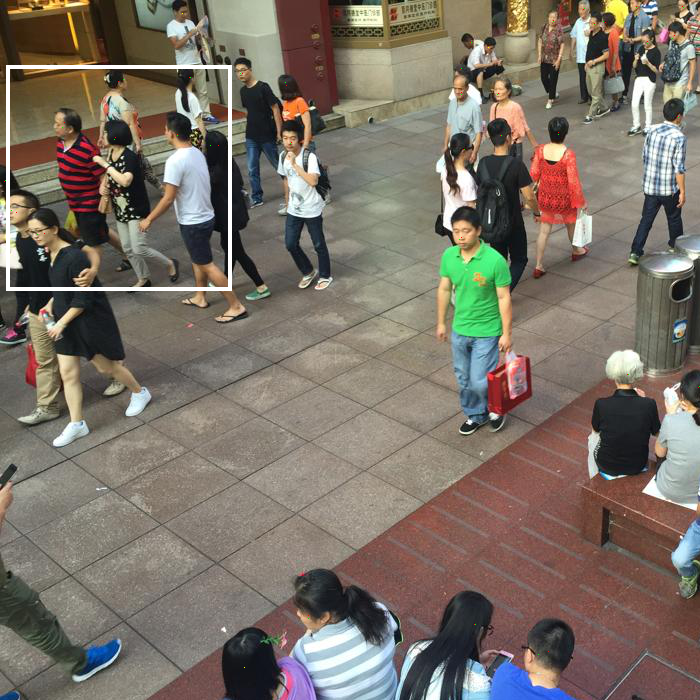}\label{Figure 3 row 1column 2}}\hspace{-2pt}
\subfloat{\includegraphics[width=.24\linewidth]{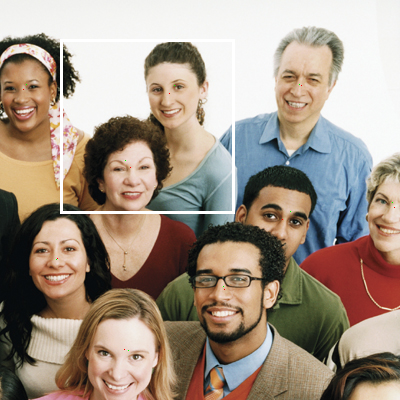}\label{Figure row 1 column 3}}\hspace{-2pt}
\subfloat{\includegraphics[width=.24\linewidth]{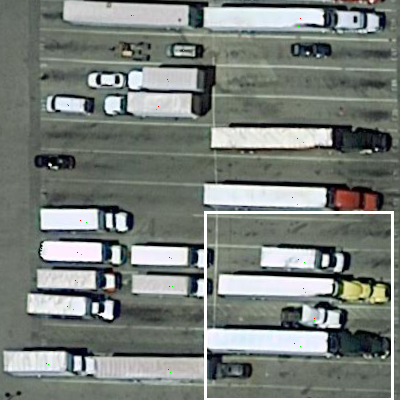}\label{Figure 3 row 2 column 3}}\hspace{-2pt}
\subfloat{\includegraphics[width=.24\linewidth]{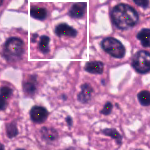}\label{Figure 3 row 3 column 8}}

\subfloat{\includegraphics[width=.24\linewidth]{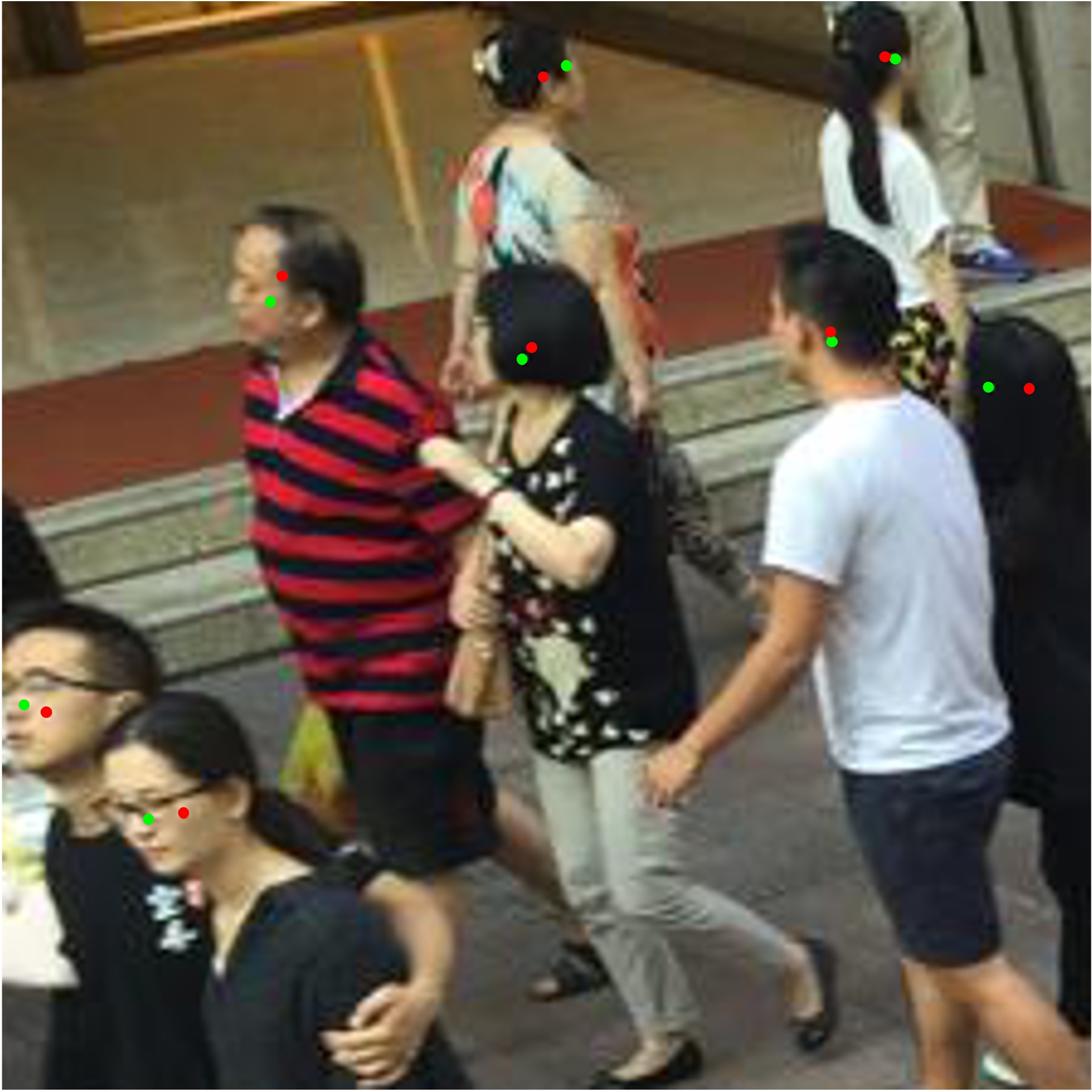}\label{Figure 3 row 1 column 4}}\hspace{-2pt}
\subfloat{\includegraphics[width=.24\linewidth]{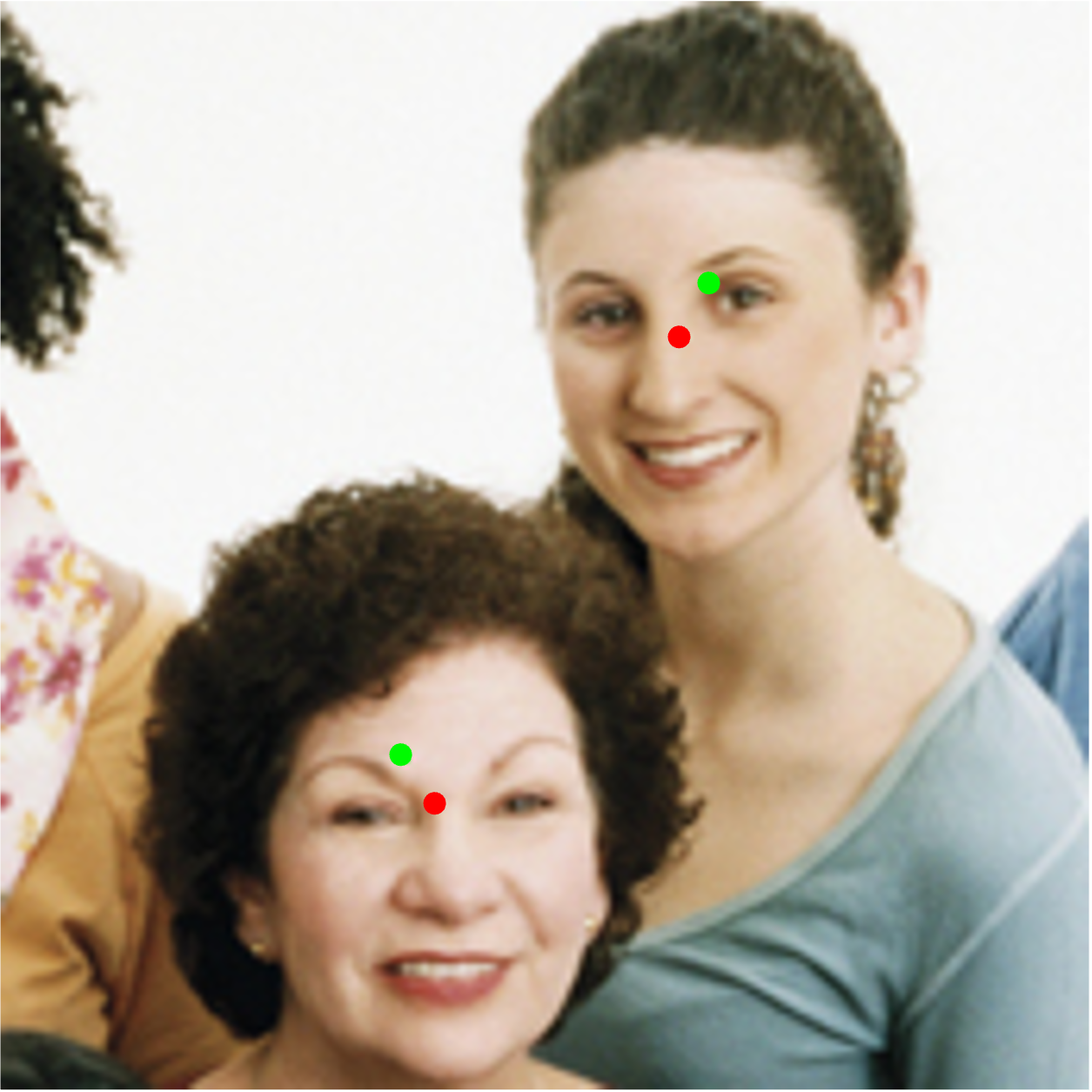}\label{Figure 3 row 2 column 2}}\hspace{-2pt}
\subfloat{\includegraphics[width=.24\linewidth]{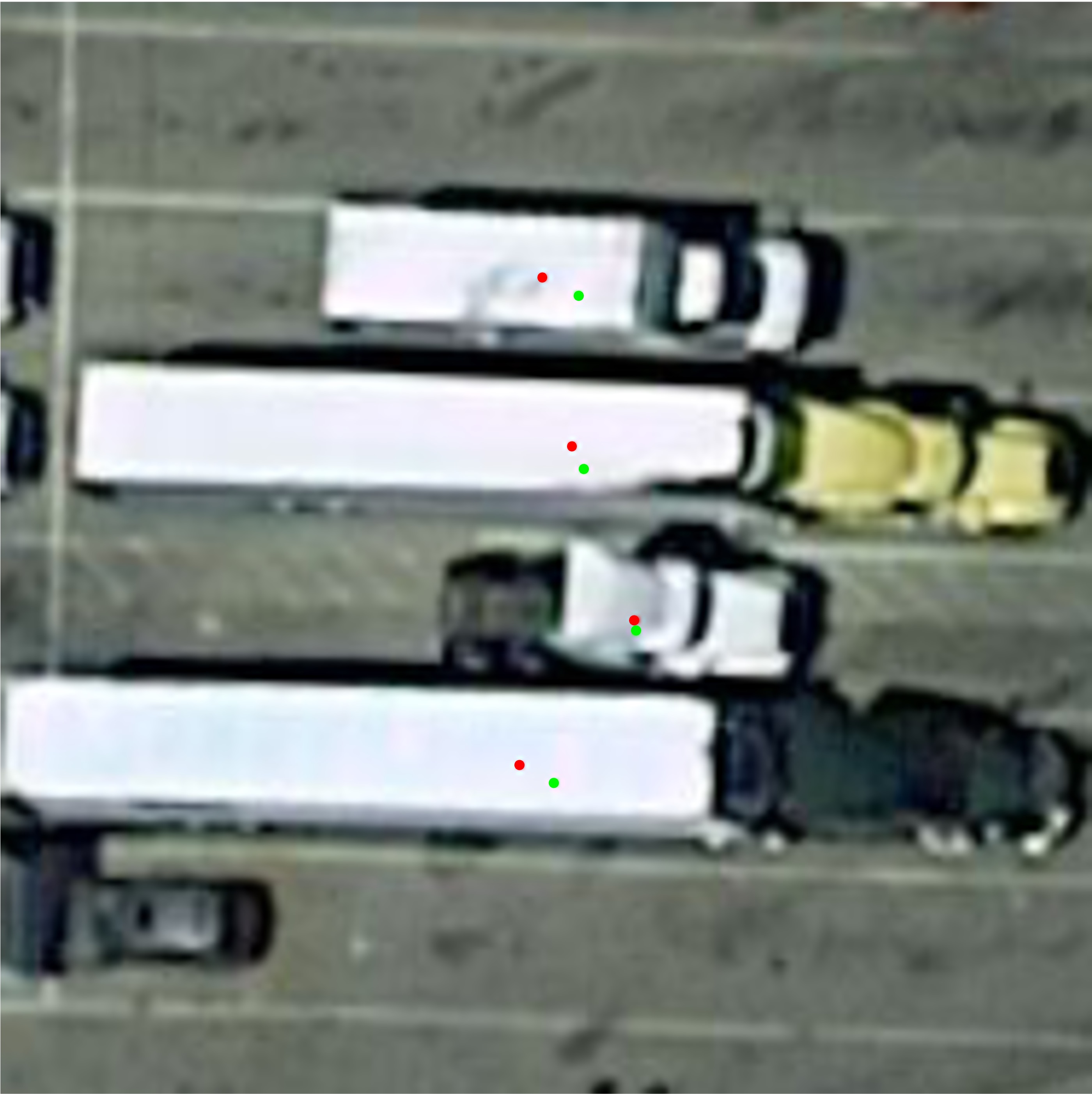}\label{Figure 3 row 2 column 4}}\hspace{-2pt}
\subfloat{\includegraphics[width=.24\linewidth]{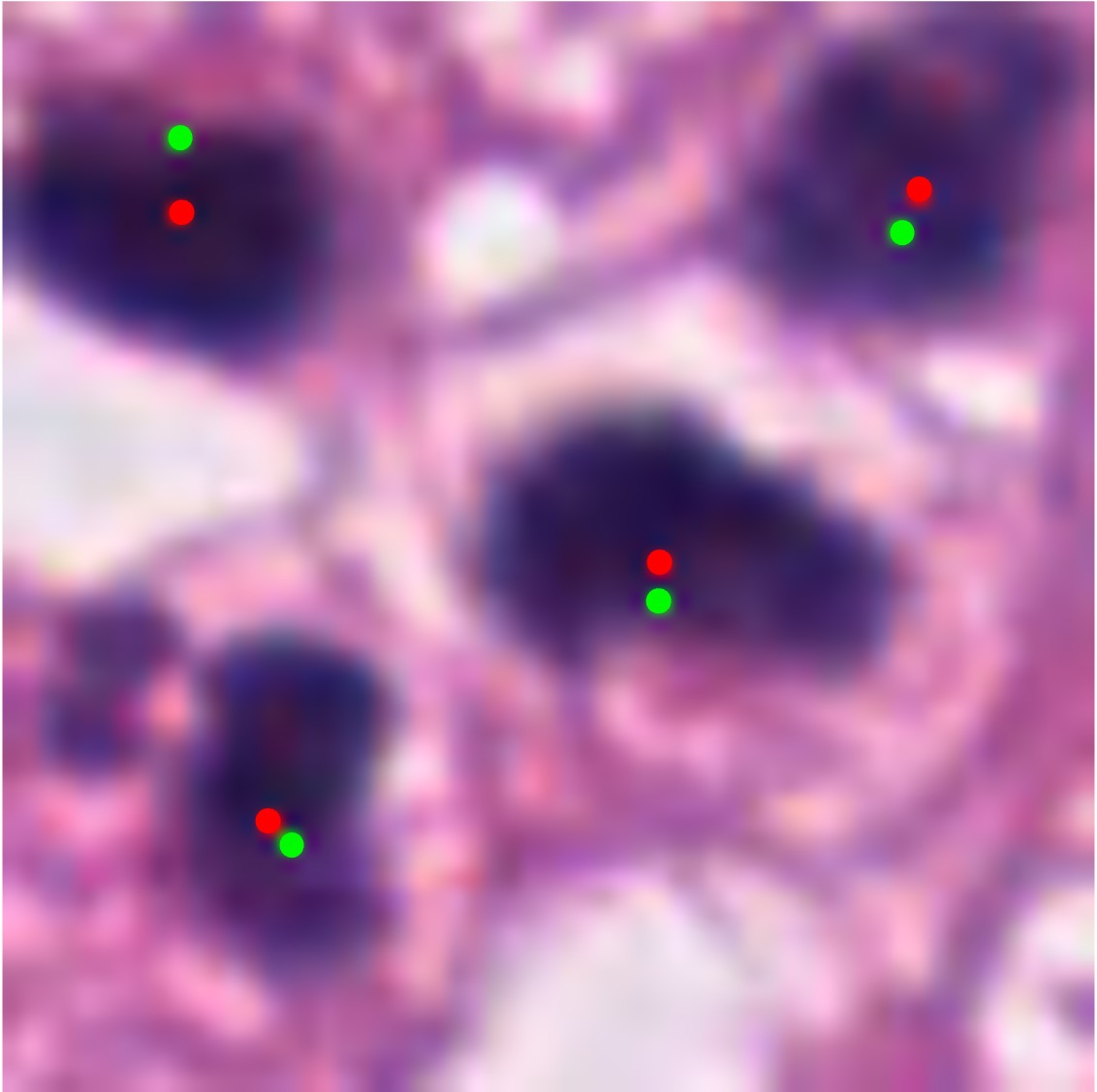}\label{Figure 3 row 3 column 1}}

\caption{Visualization of restored point annotation given by the proposed SAE. The bottom row provides a zoomed-in view within the white box of the top row. 
\textcolor{green}{Green} points: initial point annotations; \textcolor{red}{Red} points: revised point annotations; \textcolor{yellow}{Yellow} points: initial point annotation coincides with the corresponding revised point annotation.}
\label{figure: qualitative_results}
\end{figure}

\begin{table}[t]
    \centering
          \setlength{\extrarowheight}{2pt} 
	\caption{Quantitative comparison of crowd counting results on the ShanghaiTech~\cite{zhang2016single_SH_MCNN}, UCF-QNRF~\cite{idrees2018composition_UCF-QNRF}, and JHU-Crowd++~\cite{sindagi2020jhu} datasets. \& \SAEcolor{SAE} indicates training with point annotations refined by SAE. The best performance is in \textbf{boldface}. $\dag$ represents reproduced results with official codes. }
            \resizebox{\linewidth}{!}{
		\begin{tabular}{l||c c| c c| c c| c c}
            \specialrule{0.1em}{0pt}{0pt}
			\multirow{2}{*}{Method} & \multicolumn{2}{c|}{SH PartA} & \multicolumn{2}{c|}{SH PartB} & \multicolumn{2}{c|}{UCF-QNRF} & \multicolumn{2}{c}{JHU-Crowd++}  \\
            \cline{2-9}
			&  MAE$\downarrow$ & MSE$\downarrow$ & MAE$\downarrow$ & MSE$\downarrow$ & MAE$\downarrow$ & MSE$\downarrow$ & MAE$\downarrow$ & MSE$\downarrow$ \\
			 
            \specialrule{0.1em}{0pt}{0pt}
			 \rowcolor{mygray}CSRNet~\cite{li2018csrnet} \scriptsize{(CVPR'18)}                             & 68.2 & 115.0 & 10.6 & 16.0 & - & - & 85.9 & 309.2 \\
			 CAN~\cite{liu2019context_can} \scriptsize{(CVPR'19)}                                 & 62.3 & 100.0 & 7.8 & 12.2 & 107.0 & 183.0 & 100.1 & 314.0\\
			 \rowcolor{mygray}ADSCNet~\cite{Bai_2020_CVPR_ADSC} \scriptsize{(CVPR'20)} & 55.4 & 97.7 & 6.4 & 11.3 & \textbf{71.3} & 132.5 & - & - \\
			 GL~\cite{wan2021generalized_GL} \scriptsize{(CVPR'21)}                                   & 61.3 & 95.4 & 7.3 & 11.7 & 84.3 & 147.5 & 59.9 & 259.5 \\
             \rowcolor{mygray}CLTR~\cite{liang2022end_CLTR} \scriptsize{(ECCV'22)}                                & 56.9 & 95.2 & 6.5 & 10.6 & 85.8 & 141.3 & 59.5 & 240.6 \\
             NoiseCC~\cite{wan2023modeling_NoiseCC_Tpami} \scriptsize{(TPAMI’23)}                            & 61.8 & 104.3& 7.1 & 12.4 & 83.8 & 147.8 & 59.1 & 259.6 \\
             \rowcolor{mygray}NoiseCC~\cite{wan2023modeling_NoiseCC_Tpami} \& MAN \scriptsize{(TPAMI’23)}                      & 56.4 & 89.4 & 6.5 & 10.3 & 75.3 & 128.3 & 53.0 & 208.6 \\
             CrowdHat~\cite{wu2023boosting_CrowdHat} \scriptsize{(CVPR'23)}                             & 51.2 & 81.9 & 5.7 & 9.4 & 75.1 & 126.7 & 52.3 & 211.8 \\
             \rowcolor{mygray}AWCCNet~\cite{huang2023counting_AWCC} \scriptsize{(ICCV'23)}                              & 56.2 & 91.3 & - & - & 76.4 & 130.5 & 52.3 & 207.2 \\
    PET~\cite{liu2023point_PET}  \scriptsize{(ICCV'23)}                                 & 49.3 & 78.8 & 6.2 & 9.7 &79.5 & 144.3 & 58.5 & 238.0 \\
            \rowcolor{mygray}Gramformer~\cite{lin2024gramformer}  \scriptsize{(AAAI'24)}                                 & 54.7 & 87.1 & - & - &76.7 & 129.5 & 53.1 & 228.1 \\
    
            \specialrule{0.1em}{0pt}{0pt}
			 BL$^\dag$~\cite{ma2019bayesian_BL}  \scriptsize{(ICCV'19)}                                  & 62.7 & 99.5 & 7.6 & 13.00 & 87.4 & 149.6 & 74.4 & 290.0 \\
    	     \rowcolor{mygray} BL \& \SAEcolor{SAE}      & 59.5 (\SAEcolor{$\downarrow$3.2}) & 89.4 (\SAEcolor{$\downarrow$10.1}) & 6.9 (\SAEcolor{$\downarrow$0.7}) & 11.9 (\SAEcolor{$\downarrow$1.1}) & 81.6 (\SAEcolor{$\downarrow$5.8}) & 146.5 (\SAEcolor{$\downarrow$3.1}) & 60.8 (\SAEcolor{$\downarrow$13.6}) & 230.7 (\SAEcolor{$\downarrow$59.3}) \\
          \specialrule{0.04em}{0pt}{0pt}
			 NoiseCC$^\dag$~\cite{wan2020modeling_NoiseCC} \scriptsize{(NeurIPS’20)}                           & 62.1 & 100.0& 7.5 & 11.4 & 86.1 & 149.7 & 67.5 & 255.4 \\
             \rowcolor{mygray}NoiseCC \& \SAEcolor{SAE}   & 59.4 (\SAEcolor{$\downarrow$2.7}) & 91.4 (\SAEcolor{$\downarrow$8.6}) & 6.8 (\SAEcolor{$\downarrow$0.7}) & 9.8 (\SAEcolor{$\downarrow$1.6}) & 81.9 (\SAEcolor{$\downarrow$4.2}) & 134.4 (\SAEcolor{$\downarrow$15.3}) & 58.9 (\SAEcolor{$\downarrow$8.6}) & 231.8 (\SAEcolor{$\downarrow$23.6}) \\
             \specialrule{0.04em}{0pt}{0pt}
             RSI-ResNet50$^\dag$~\cite{rsi_cheng2022rethinking} \scriptsize{(CVPR'22)}                 & 54.4 & 89.0 & 6.6 & 9.8  & 81.2 & 152.0 & 58.8 & 245.1 \\
    	   \rowcolor{mygray} RSI-ResNet50 \& \SAEcolor{SAE}                                            & 52.4 (\SAEcolor{$\downarrow$2.0}) & 85.4 (\SAEcolor{$\downarrow$3.6}) & 6.0 (\SAEcolor{$\downarrow$0.6}) & 9.2 (\SAEcolor{$\downarrow$0.6})  & 77.5 (\SAEcolor{$\downarrow$3.7}) & 146.2 (\SAEcolor{$\downarrow$5.8}) & 54.9 (\SAEcolor{$\downarrow$3.9}) & 240.2 (\SAEcolor{$\downarrow$4.9}) \\
             \specialrule{0.1em}{0pt}{0pt}
			 P2PNet$^\dag$~\cite{song2021rethinking_P2P} \scriptsize{(ICCV'21)}                               & 52.8 & 85.8 & 6.5 & 10.9 & 91.7 & 157.0 & 66.8 & 259.5 \\
             \rowcolor{mygray}P2PNet \& \SAEcolor{SAE}   & \textbf{48.2} (\SAEcolor{$\downarrow$4.6}) & \textbf{76.1} (\SAEcolor{$\downarrow$9.7}) & 6.2 (\SAEcolor{$\downarrow$0.3}) & 10.0 (\SAEcolor{$\downarrow$0.9})  & 85.9 (\SAEcolor{$\downarrow$5.8}) & 148.6 (\SAEcolor{$\downarrow$8.4}) & 62.4 (\SAEcolor{$\downarrow$4.4}) & 253.7 (\SAEcolor{$\downarrow$5.8}) \\
             \specialrule{0.04em}{0pt}{0pt}
			 MAN$^\dag$~\cite{lin2022boosting_MAN} \scriptsize{(CVPR’22)}        & 55.6 & 93.2 & 7.1 & 10.5 & 77.5 & 132.7 & 53.2 & 219.9 \\
			 \rowcolor{mygray}MAN \& \SAEcolor{SAE}        & 52.2 (\SAEcolor{$\downarrow$3.4}) & 81.9 (\SAEcolor{$\downarrow$11.3}) & \textbf{5.4} (\SAEcolor{$\downarrow$1.7}) & \textbf{7.0} (\SAEcolor{$\downarrow$3.5})  & 74.2 (\SAEcolor{$\downarrow$3.3}) & 128.2 (\SAEcolor{$\downarrow$4.5}) &\textbf{49.7} (\SAEcolor{$\downarrow$3.5}) & \textbf{204.4} (\SAEcolor{$\downarrow$15.5}) \\
            \specialrule{0.04em}{0pt}{0pt}
    	     STEERER$^\dag$~\cite{han2023steerer_STEERER} \scriptsize{(ICCV’23)}                              & 56.5 & 89.8 & 7.1 & 10.7 & 74.1  & 129.5  & 55.8  & 223.2 \\
			 \rowcolor{mygray}STEERER \& \SAEcolor{SAE}     & 54.3 (\SAEcolor{$\downarrow$2.2}) & 84.5 (\SAEcolor{$\downarrow$5.3}) & 6.4 (\SAEcolor{$\downarrow$0.7}) & 10.1 (\SAEcolor{$\downarrow$0.6}) & 71.4 (\SAEcolor{$\downarrow$2.7}) & \textbf{125.1} (\SAEcolor{$\downarrow$4.4}) & 52.9 (\SAEcolor{$\downarrow$2.9}) & 216.4 (\SAEcolor{$\downarrow$6.8}) \\
            \specialrule{0.1em}{0pt}{0pt}
	    \end{tabular}}
	\label{table: crowd_counting}
\end{table}

\subsection{Experimental results}
\label{subsec: experimental results}
We conduct extensive object counting experiments on eleven publicly available datasets across three distinct domains: crowd counting, remote sensing object counting, and cell counting. 
Firstly, Fig.~\ref{figure: qualitative_results} presents some qualitative results of restored point annotation given by our SAE, illustrating the SAE's effectiveness in enhancing the consistency of point annotations for various object types. To further quantify SAE's effectiveness, we conduct quantitative comparisons in terms of object counting accuracy by training some existing counting methods (including both density-map-based and localization-based approaches) with initial annotations and restored annotations by SAE, respectively. The detailed quantitative comparison is given in the following.

\medskip
\noindent
\textbf{Crowd counting:} We mainly evaluate SAE on two types of counting methods. 

\noindent
\textit{\textbf{1)} Comparison with state-of-the-art methods.}
We conduct a comprehensive evaluation of the SAE-refined point annotations compared with the initial ones, employing a variety of SOTA density-map-based and localization-based crowd counting models.
The quantitative results, depicted in Tab.~\ref{table: crowd_counting}, demonstrate the effectiveness of SAE on various baseline methods.
A notable improvement is observed when applied to P2PNet~\cite{song2021rethinking_P2P}, where SAE achieves a reduction of 5.8 in MAE and 8.4 in MSE on UCF-QNRF dataset. 
Remarkably, the integration of SAE leads to the establishment of new state-of-the-art results in three of the four assessed datasets: SH PartA~\cite{zhang2016single_SH_MCNN}, SH PartB~\cite{zhang2016single_SH_MCNN}, and JHU++~\cite{sindagi2020jhu}.
On average, SAE contributes to a reduction of 3.0 in MAE and 6.4 in MSE compared to the results of three baseline methods (P2PNet~\cite{song2021rethinking_P2P}, MAN~\cite{lin2022boosting_MAN}, and STEERER~\cite{han2023steerer_STEERER}) on the four extensively used datasets.

\begin{table}[t]
    \centering
      \setlength{\extrarowheight}{2pt} 
	\caption{Quantitative comparison of remote sensing object counting results on RSOC datasets~\cite{gao2020counting_ASPD_RSOC}. \& \SAEcolor{SAE} indicates training with point annotations refined by SAE. The best performance is in \textbf{boldface}. $\dag$ represents reproduced results with official codes. }
  \resizebox{\linewidth}{!}{
		\begin{tabular}{l || c c| c c| c c| c c}
			\specialrule{0.1em}{0pt}{0pt}
			\multirow{2}{*}{Method} & \multicolumn{2}{c|}{RSOC\_Building} & \multicolumn{2}{c|}{RSOC\_Small-vehicle} & \multicolumn{2}{c|}{RSOC\_Large-vehicle} & \multicolumn{2}{c}{RSOC\_Ship}  \\
            \cline{2-9}
			& MAE$\downarrow$ & MSE$\downarrow$ & MAE$\downarrow$ & MSE$\downarrow$ & MAE$\downarrow$ & MSE$\downarrow$ & MAE$\downarrow$ & MSE$\downarrow$ \\
			 
			 \specialrule{0.1em}{0pt}{0pt}
             \rowcolor{mygray}MCFA~\cite{duan2021distillation_MCFA} \scriptsize{(TGRS'21)}                                     &7.9   &11.8 &238.5  & 625.9 & 12.9 & 20.3 & 50.5 & 65.2   \\
             ADMAL~\cite{ding2022object_ADMAL} \scriptsize{(TGRS'22) }                       &\textbf{5.6}   &\textbf{7.7}  &115.6  & 210.8 & 11.7 & 17.3 & 45.1 & 64.8   \\
             \rowcolor{mygray}eFreeNet~\cite{huang2023remote_eFree} \scriptsize{(TGRS'23)}                                 &\textbf{5.6}   &\textbf{7.7}  &195.9  & 463.6 & 14.6 & 19.8 & 65.3 & 85.5   \\
            LMSFFNet~\cite{yi2023lightweight_LMSFFNet} \scriptsize{(TGRS'23)}   &6.3   &9.4  &141.7  & 273.0 & 12.7 & 27.1 & 49.5 & 70.0   \\
            \rowcolor{mygray}DOPNet~\cite{cui2024dopnet} \scriptsize{(TGRS'24)}   &-   &-  &62.4  & \textbf{167.8} & 12.5 & 20.1 & - & -   \\

			 \specialrule{0.1em}{0pt}{0pt}
			 BL$^\dag$~\cite{ma2019bayesian_BL} \scriptsize{(ICCV'19)}                                         & 11.5 & 16.3 & 173.0 & 477.8 & 12.3 & 24.2 & 58.8 & 195.5 \\
    	     \rowcolor{mygray} BL \& \SAEcolor{SAE}                                & 11.1 (\SAEcolor{$\downarrow$0.4}) & 15.9 (\SAEcolor{$\downarrow$0.4}) & 130.6 (\SAEcolor{$\downarrow$42.4})  & 341.4 (\SAEcolor{$\downarrow$136.4}) & 8.5 (\SAEcolor{$\downarrow$3.8})  & 14.6 (\SAEcolor{$\downarrow$9.6}) & 51.0 (\SAEcolor{$\downarrow$7.8}) & 73.6 (\SAEcolor{$\downarrow$121.9}) \\
            \specialrule{0.04em}{0pt}{0pt}
			 NoiseCC$^\dag$~\cite{wan2020modeling_NoiseCC} \scriptsize{(NeurIPS’20)}                                  & 7.8  & 11.3 & 168.8 & 529.1 & 15.5 & 31.4 & 53.0 & 72.1  \\
             \rowcolor{mygray} NoiseCC \& \SAEcolor{SAE}         & 7.3 (\SAEcolor{$\downarrow$0.5})  & 10.2 (\SAEcolor{$\downarrow$1.1}) & 149.7 (\SAEcolor{$\downarrow$19.1}) & 389.4 (\SAEcolor{$\downarrow$139.7}) & 14.6 (\SAEcolor{$\downarrow$0.9}) & 30.8 (\SAEcolor{$\downarrow$0.6}) & 46.3 (\SAEcolor{$\downarrow$6.7}) & 69.0 (\SAEcolor{$\downarrow$3.1})  \\
             \specialrule{0.1em}{0pt}{0pt}
			 P2PNet$^\dag$~\cite{song2021rethinking_P2P} \scriptsize{(ICCV'21)}                                      & 6.3  & 9.1  & 63.1 & 198.3 & 8.1  & 13.2 & 28.2 & 42.6  \\
             \rowcolor{mygray} P2PNet \& \SAEcolor{SAE}            & 5.7 (\SAEcolor{$\downarrow$0.6})  & 8.1 (\SAEcolor{$\downarrow$1.0})  & \textbf{53.3} (\SAEcolor{$\downarrow$9.8})  & 170.4 (\SAEcolor{$\downarrow$27.9}) & \textbf{7.0} (\SAEcolor{$\downarrow$1.1})  & \textbf{11.6} (\SAEcolor{$\downarrow$1.6}) & \textbf{25.0} (\SAEcolor{$\downarrow$3.2}) & \textbf{39.5} (\SAEcolor{$\downarrow$3.1})  \\
             \specialrule{0.04em}{0pt}{0pt}
             ASPDNet$^\dag$~\cite{gao2020counting_ASPD_RSOC} \scriptsize{(TGRS'20)}                                 & 7.6  & 11.5 &252.6  & 718.5 & 19.7 & 27.8 & 81.8 & 110.3 \\
             \rowcolor{mygray} ASPDNet \& \SAEcolor{SAE}         & 6.7 (\SAEcolor{$\downarrow$0.9})  & 10.6 (\SAEcolor{$\downarrow$0.9}) &182.0 (\SAEcolor{$\downarrow$70.6})  & 353.4 (\SAEcolor{$\downarrow$365.1}) & 16.9 (\SAEcolor{$\downarrow$2.8}) & 24.4 (\SAEcolor{$\downarrow$3.4}) & 75.1 (\SAEcolor{$\downarrow$6.7}) & 97.0 (\SAEcolor{$\downarrow$13.3})  \\
             \specialrule{0.04em}{0pt}{0pt}
             PSGCNet$^\dag$~\cite{gao2022psgcnet} \scriptsize{(TGRS'22)}                                 &7.4   &11.1  &112.1  & 289.6 & 11.8 & 16.4 & 39.5 & 68.5   \\
             \rowcolor{mygray} PSGCNet \& \SAEcolor{SAE}         &6.6 (\SAEcolor{$\downarrow$0.8})   &10.2 (\SAEcolor{$\downarrow$0.9})  &104.9 (\SAEcolor{$\downarrow$7.2})  & 227.2 (\SAEcolor{$\downarrow$62.4}) & 9.7 (\SAEcolor{$\downarrow$2.1})  & 14.0 (\SAEcolor{$\downarrow$2.4}) & 34.6 (\SAEcolor{$\downarrow$4.9}) & 62.8 (\SAEcolor{$\downarrow$5.7})   \\

    	   \specialrule{0.1em}{0pt}{0pt}
	    \end{tabular}}
	\label{table: remote_sensing_object_couning}
\end{table}

\smallskip
\noindent
\textit{\textbf{2)} Effectiveness on anti-noise methods.}
Further investigations are conducted to assess SAE's compatibility with those anti-noise approaches: BL~\cite{ma2019bayesian_BL}, NoiseCC~\cite{wan2020modeling_NoiseCC}, and RSI~\cite{rsi_cheng2022rethinking}.
While these methods primarily focus on improving noise resistance, they do not inherently improve the quality of the initial point annotations.
As depicted in Tab.~\ref{table: crowd_counting}, SAE further steadily improves these anti-noise frameworks, yielding an average enhancement of 4.1 in MAE and 11.5 in MSE across the four diverse crowd counting datasets.

\medskip
\noindent
\textbf{Remote sensing object counting:} We mainly evaluate SAE on two types of counting methods. 

\noindent
\textit{\textbf{1)} Comparison with state-of-the-art methods.}
In contrast to crowd counting datasets, remote sensing object counting task presents a different challenge with their diversity, encompassing four distinct types of objects.
We apply SAE to several state-of-the-art remote sensing object counting methods. 
The quantitative results, detailed in Tab.~\ref{table: remote_sensing_object_couning}, clearly demonstrate the effective performance of SAE.
It not only consistently outperforms baseline methods but also sets new state-of-the-art results in three of the four datasets: RSOC\_Small-vehicle, RSOC\_Large-vehicle, and RSOC\_Ship~\cite{gao2020counting_ASPD_RSOC}.
Notably, SAE achieves, on average, an improvement of 9.2 in MAE and 40.6 in MSE over the three baseline models (P2PNet~\cite{song2021rethinking_P2P}, ASPDNet~\cite{gao2020counting_ASPD_RSOC}, and PSGCNet~\cite{gao2022psgcnet}) across these four datasets.

\smallskip
\noindent
\textit{\textbf{2)} Effectiveness on anti-noise methods.}
In addition to state-of-the-art approaches, we also examine the effectiveness of our SAE on two anti-noise methods, BL~\cite{ma2019bayesian_BL} and NoiseCC~\cite{wan2020modeling_NoiseCC}.
These anti-noise methods aim to enhance the model's noise resilience, primarily through adjustments in the loss functions.
The comparative results shown in Tab.~\ref{table: remote_sensing_object_couning} reveal that SAE further augments the performance for these anti-noise methods, resulting in an average improvement of 10.2 in MAE and 51.6 in MSE across the four datasets.

\medskip
\noindent
\textbf{Cell counting:} We mainly evaluate SAE on two types of counting methods. 

\noindent
\textit{\textbf{1)} Comparison with state-of-the-art methods.}
The proposed SAE method extends its utility beyond natural image datasets. We also demonstrate its ability to resolve annotation inconsistency challenges encountered in cell counting. As detailed in Tab.~\ref{table: cell counting}, the proposed SAE consistently improves cell counting accuracy. In particular,
we establish new state-of-the-art records across all the three cell counting datasets.
Notably, with SAU~\cite{guo2021sau} already achieving excellent results on DCC~\cite{paper_DCC} with an MAE of 3.0 and an MSE of 4.8, using the restored point annotation by SAE further enhances the performance, reducing the MAE to 2.6 and MSE to 3.4, respectively.

\smallskip
\noindent
\textit{\textbf{2)} Effectiveness on anti-noise methods.}
Additionally, we also explore SAE's universal applicability in conjunction with two established anti-noise methods: BL~\cite{ma2019bayesian_BL} and NoiseCC~\cite{wan2020modeling_NoiseCC}.
These two methods focus on improving noise robustness by introducing novel loss functions. As shown in Tab.~\ref{table: cell counting}, the proposed SAE consistently improves their performance, which further validates the effectiveness and universality of the proposed SAE.

\begin{table}[t]
    \centering
	\caption{Quantitative comparison of cell counting results on MBM~\cite{paper_MBM}, ADI~\cite{paper_ADI}, and DCC~\cite{paper_DCC}. \& \SAEcolor{SAE} indicates training with point annotations refined by SAE. The best performance is in \textbf{boldface}. $\dag$ represents reproduced results with official codes. }
 \setlength{\extrarowheight}{2pt} 

        \resizebox{0.9\linewidth}{!}{
		\begin{tabular}{l ||c c| c c| c c}
			\specialrule{0.1em}{0pt}{0pt}
			\multirow{2}{*}{Method} & \multicolumn{2}{c|}{MBM} & \multicolumn{2}{c|}{ADI}  & \multicolumn{2}{c}{DCC}  \\
            \cline{2-7}
			&  MAE$\downarrow$ & MSE$\downarrow$ & MAE$\downarrow$ & MSE$\downarrow$ & MAE$\downarrow$ & MSE$\downarrow$ \\
			 
			 \specialrule{0.1em}{0pt}{0pt}
			 BL$^\dag$~\cite{ma2019bayesian_BL} \scriptsize{(ICCV'19)}                                                        &8.4   &10.3   &13.7 &18.8 &3.3   &5.3   \\
    	      \rowcolor{mygray}BL \& \SAEcolor{SAE}                                                                                 &6.3 (\SAEcolor{$\downarrow2.1$})   &8.8 (\SAEcolor{$\downarrow1.5$})    &13.2 (\SAEcolor{$\downarrow0.5$}) &18.2 (\SAEcolor{$\downarrow0.6$}) &2.9 (\SAEcolor{$\downarrow0.4$})   &4.9 (\SAEcolor{$\downarrow0.4$})   \\
           \specialrule{0.04em}{0pt}{0pt}
			NoiseCC$^\dag$~\cite{wan2020modeling_NoiseCC} \scriptsize{(NeurIPS’20)}                                          &7.2   &9.3    &14.0 &18.8 &3.8   &5.6   \\
             \rowcolor{mygray}NoiseCC \& \SAEcolor{SAE}                         &6.0 (\SAEcolor{$\downarrow1.2$})   &8.6 (\SAEcolor{$\downarrow0.7$})    &13.1 (\SAEcolor{$\downarrow0.9$}) &17.9 (\SAEcolor{$\downarrow0.9$}) &3.4 (\SAEcolor{$\downarrow0.4$})   &5.1 (\SAEcolor{$\downarrow0.5$})   \\
             \specialrule{0.1em}{0pt}{0pt}
			P2PNet$^\dag$~\cite{song2021rethinking_P2P} \scriptsize{(ICCV'21)}                                                    &5.7   &8.0    &12.4 &17.1 &3.7   &6.0   \\
              \rowcolor{mygray}P2PNet \& \SAEcolor{SAE}                              &4.4 (\SAEcolor{$\downarrow1.3$})   &6.5 (\SAEcolor{$\downarrow1.5$})    &11.9 (\SAEcolor{$\downarrow0.5$}) &16.2 (\SAEcolor{$\downarrow0.9$}) &3.2 (\SAEcolor{$\downarrow0.5$})   &5.6 (\SAEcolor{$\downarrow0.4$})   \\
              \specialrule{0.04em}{0pt}{0pt}
             Chen et al.$^\dag$~\cite{chen_direction_field} \scriptsize{(JBHI'21)}                                            &5.6   &8.3    &12.7 &19.6 &5.7   &7.6   \\
             \rowcolor{mygray}Chen et al. \& \SAEcolor{SAE}                      &4.5 (\SAEcolor{$\downarrow1.1$})   &7.5 (\SAEcolor{$\downarrow0.8$})    &11.4 (\SAEcolor{$\downarrow1.3$}) &15.8 (\SAEcolor{$\downarrow3.8$}) &4.7 (\SAEcolor{$\downarrow1.0$})   &7.1 (\SAEcolor{$\downarrow0.5$})   \\
             \specialrule{0.04em}{0pt}{0pt}
             SAUNet$^\dag$~\cite{guo2021sau} \scriptsize{(IEEE ACM T COMPUT BI'21)}                                       &5.7   &7.7    &14.3 &18.5 &3.0   &4.8   \\
             \rowcolor{mygray}SAUNet \& \SAEcolor{SAE}                            &\textbf{4.2} (\SAEcolor{$\downarrow1.5$})   &\textbf{5.8} (\SAEcolor{$\downarrow1.9$})    &\textbf{11.2} (\SAEcolor{$\downarrow3.1$}) &\textbf{15.1} (\SAEcolor{$\downarrow3.4$}) &\textbf{2.6} (\SAEcolor{$\downarrow0.4$})   &\textbf{3.4} (\SAEcolor{$\downarrow1.4$})   \\

    	\specialrule{0.1em}{0pt}{0pt}
	    \end{tabular}}
	\label{table: cell counting}
\end{table}

\subsection{Robustness to extra point annotation noise}
\label{subsec: robustness}
We also evaluate the robustness of our SAE to point annotation noise of different degrees. For that, 
Firstly, we emulate real-world human annotation noise errors by introducing varying offsets to each annotation point $p$. The magnitude of offsets are set to $0.2\times r$, $0.4\times r$, $0.6\times r$, and $0.8\times r$ pixels, where $r$ is defined in Eq.~\eqref{eq: radius_with_distance_and_limitation}. 
Secondly, we train two anti-noise crowd counting methods: BL~\cite{ma2019bayesian_BL} and NoiseCC~\cite{wan2020modeling_NoiseCC} on these noisy point annotations. We then apply our SAE to restore the noisy point annotations, and train a new BL on our restored point annotations. These experiments are conducted on the UCF-QNRF~\cite{idrees2018composition_UCF-QNRF} dataset. 

\begin{figure}[t]
  \centering
  \subfloat{\includegraphics[width=.6\linewidth]{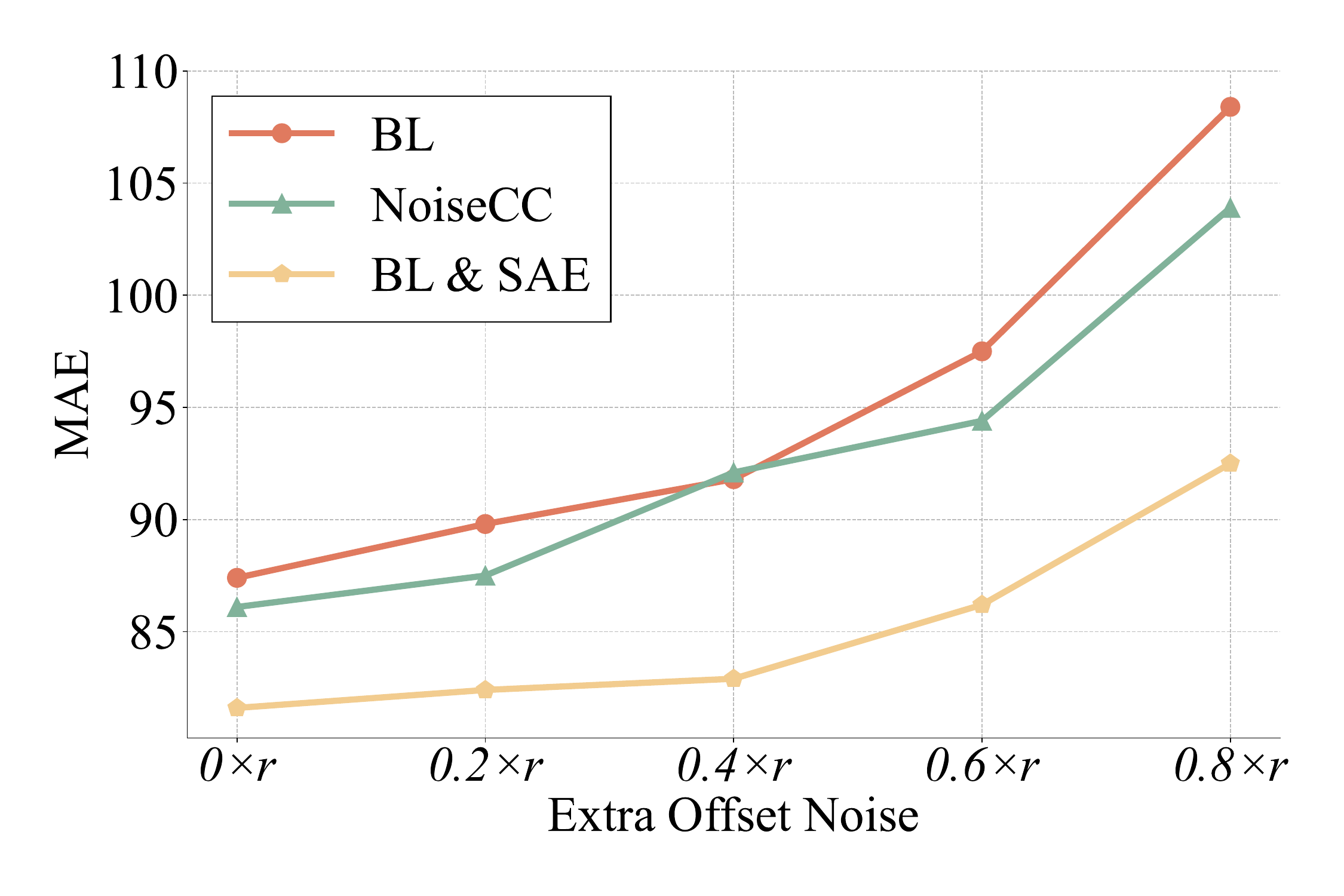}\label{Figure robustness to noisy point annotations mae}}
  \caption{Evaluation of counting robustness to different levels of synthetic offset noise on the UCF-QNRF~\cite{idrees2018composition_UCF-QNRF} dataset. We shift the original point annotation by different proportions of radius $r$ (defined in Eq.~\eqref{eq: radius_with_distance_and_limitation}) in random directions to synthesize annotation noise of different levels. }
  \label{figure: robustness to noisy point annotations}
\end{figure}

As shown in Fig.~\ref{figure: robustness to noisy point annotations}, increasing noise levels of annotation poses challenges to all methods in maintaining counting accuracy. Remarkably, SAE consistently outperforms the other two techniques and is less affected by increasing noise levels (in particular for extra noise $\leq 0.4\times r$), achieving lower MAE. These findings confirm SAE's effectiveness and robustness in handling noisy and imperfect point annotation, a common occurrence in real-world counting tasks.

\subsection{Ablation study on Hyper-Parameter $\alpha$ in Eq.~\eqref{eq: radius_with_distance_and_limitation}}
\label{subsec: abation_alpha}
The proposed SAE mainly involves one hyper-parameter $\alpha$, which is involved in Eq.~\eqref{eq: radius_with_distance_and_limitation} and plays a crucial role in defining the sampling area for shifted point generation. To evaluate the impact of the hyper-parameter $\alpha$ on our model's performance, we conduct an ablation study with $\alpha$ values set to 0.3, 0.4, 0.5, and 0.6. This study utilizes BL~\cite{ma2019bayesian_BL} as the baseline counting method, and covers three diverse datasets: UCF-QNRF~\cite{idrees2018composition_UCF-QNRF}, Remote\_small-vehicle~\cite{gao2020counting_ASPD_RSOC}, and MBM~\cite{paper_MBM}. 

The results, presented in Tab.~\ref{table: ablation_alpha}, show how different $\alpha$ values influence counting accuracy. Generally, in configurations that avoid overlap, we observe performance improvements compared to the baseline method. For $\alpha$ set to 0.6, the overlapping sampling ranges for different objects introduce confusion in the training process, leading to performance that falls short of the baseline. Indeed, a smaller value of $\alpha$ leads to a more constrained sampling range, which may reduce the SAE model's effectiveness in refining annotations with large deviations from their general positions. In contrast, a larger $\alpha$ tends to include more background regions within the sampling range. Notably, when the sampling areas for different annotation points overlap, it leads to ambiguity. This overlap challenges SAE's ability to distinguish between regions associated with distinct objects. Except for this ablation study, we set $\alpha$ to 0.4 for our SAE in all experiments.

\begin{table}[t]
\centering
  \caption{Quantitative results of BL~\cite{ma2019bayesian_BL} (on UCF-QNRF~\cite{idrees2018composition_UCF-QNRF}, RSOC\_Small-vehicle (RSV)~\cite{gao2020counting_ASPD_RSOC}, and MBM~\cite{paper_MBM} dataset) trained with point annotations refined by SAE with different values of $\alpha$ involved in Eq.~\eqref{eq: radius_with_distance_and_limitation}. The best performance is in boldface.
  }
    \resizebox{0.6\linewidth}{!}{
  \begin{tabular}{l|| c c| c c| c c}
    \specialrule{0.1em}{0pt}{0pt}
    \multirow{2}{*}{$\alpha$} & \multicolumn{2}{c|}{UCF-QNRF} & \multicolumn{2}{c|}{RSV} & \multicolumn{2}{c}{MBM} \\
    \cline{2-7} 
			&  MAE$\downarrow$ & MSE$\downarrow$ & MAE$\downarrow$ & MSE$\downarrow$ & MAE$\downarrow$ & MSE$\downarrow$\\
    
    \specialrule{0.1em}{0pt}{0pt}
    \rowcolor{mygray}Baseline         &    \makebox[0.02\textwidth]{} 87.4  \makebox[0.02\textwidth]{}   &  \makebox[0.02\textwidth]{} 149.6  \makebox[0.02\textwidth]{}  &  \makebox[0.02\textwidth]{} 173.0  \makebox[0.02\textwidth]{} &  \makebox[0.02\textwidth]{} 477.8  \makebox[0.02\textwidth]{} &   \makebox[0.02\textwidth]{} 8.4  \makebox[0.02\textwidth]{}   &  \makebox[0.02\textwidth]{} 10.3  \makebox[0.02\textwidth]{}     \\
    \specialrule{0.1em}{0pt}{0pt}
    0.3              &   86.5    & 148.3   & 149.7  & 389.4 &     7.3   & 9.2     \\
    \rowcolor{mygray}0.4              &   \textbf{81.6}   &146.5   & \textbf{130.6}   & \textbf{341.4}  &   \textbf{6.3}   & 8.8     \\
    0.5              &   82.4    &\textbf{146.3}    & 138.2   & 364.8  &     6.9    & \textbf{8.6}    \\ 
    \rowcolor{mygray}0.6              &   91.0     &158.9 & 190.6  & 516.5 &   8.3    & 10.8    \\
  \specialrule{0.1em}{0pt}{0pt}

\end{tabular}}
\label{table: ablation_alpha}
\end{table}

\subsection{Limitation}
\label{subsec: limitation}
As described in Sec.~\ref{subsec: methodology_generation}, the proposed SAE approximates the radius of the sampling region based on the spatial distribution of objects.
While this strategy is functional and simple, there could be more effective strategies for determining the optimal radius, such as depth estimation.
Nevertheless, these alternatives might complicate the proposed method and compromise its universality.

%% file: paper_sections/5_conclusion.tex
\section{Conclusion}
\label{sec:conclusion}

In this paper, we focus on the inconsistency problem of point annotations in object counting tasks. This is often caused by the inevitable subjective nature of annotators. To mitigate this issue, we propose the novel Shifted Autoencoders (SAE) to revise the initial point annotations. Similar to MAE which uses general knowledge to reconstruct masked areas, resulting in specific-detail-free image reconstruction, our SAE leverages general positional knowledge to restore shifted annotations, yielding specific-offset-noise-free point restoration. Using such restored consistent point annotation to train the counting model improves the counting accuracy. Extensive experiments on eleven datasets from three different object counting tasks verify the effectiveness of the proposed SAE. Besides, based on the proposed SAE, we set new state-of-the-art results on nine of the eleven datasets. We hope that this work could shed light on the research direction of directly refining the annotation in point-based vision tasks.

%% file: paper_sections/X_supple.tex
\renewcommand{\thefigure}{S-\arabic{figure}}
\renewcommand{\thetable}{S-\arabic{table}}
\renewcommand{\thesection}{S-\arabic{section}}

In this supplementary submission, we first conduct a toy experiment to quantify the enhanced consistency of the point annotations restored by SAE in Sec.~\ref{supple_sec: toy experiment}. Then, we validate the effectiveness of further using trained SAE as the pretrained model for counting task (similar to Masked Autoencoders (MAE)~\cite{he2022mae} for downstream tasks) in Sec.~\ref{supple_sec: pretraining}. Lastly, some additional visualization results of the point annotations restored by SAE are present in Sec.~\ref{supple_sec: visualization}.

\section{Toy experiment}
\label{supple_sec: toy experiment}
In the main paper, we demonstrate that the point annotations revised by SAE are more consistent than the initial manual point annotations via some qualitative visualizations and effectiveness for final counting on eleven widely used datasets. 
Here we design a toy experiment to further validate the effectiveness of SAE by quantifying the consistency of the point annotations restored by SAE.

\begin{figure}[h]
  \centering
  \includegraphics[width=0.7\linewidth]{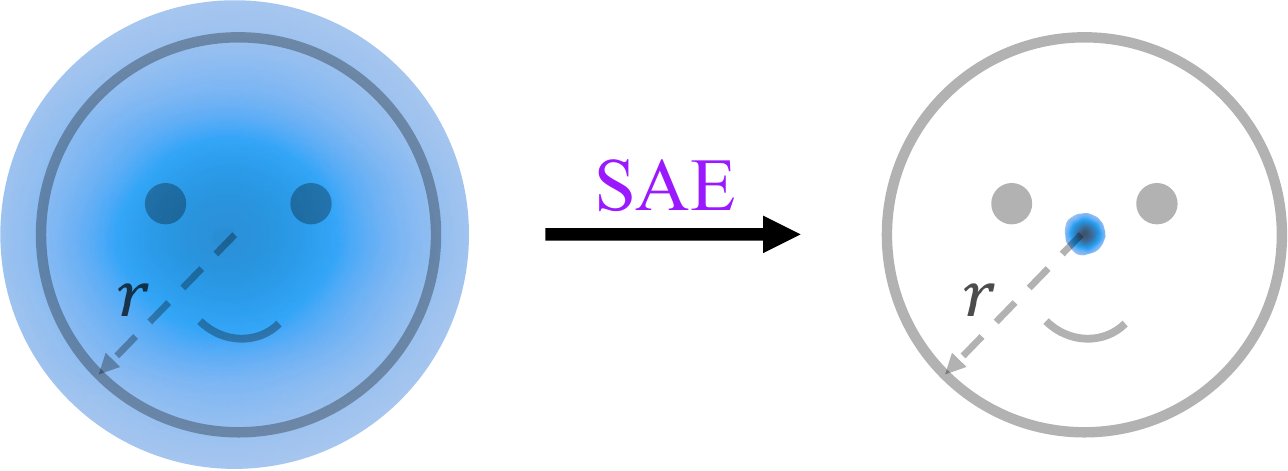}
  \caption{The distributions of point annotations relative to the stick figure's heads before and after SAE restoration. Darker blue hues signify higher distribution density. }
  \label{fig: suppple_distribution}
\end{figure}

Specifically, we generate many images of identical stick figure heads (a circle shape with radius $r$) as training images. Since the relative spatial distribution of manual point annotations \textit{w.r.t.} corresponding objects often roughly obeys a Gaussian function~\cite{wan2023modeling_NoiseCC_Tpami}, we simulate the manual point annotations with random offset noise for each stick figure head by randomly sampling from a two-dimensional Gaussian distribution, which is centered at the center of the head circle. 
The distribution of simulated point annotations for the stick figure heads is illustrated in the left part of Fig.~\ref{fig: suppple_distribution}. Some annotations may even locate outside of the head region. This phenomenon also occurs in actual object counting datasets, which is shown by the green point within the white boxed area in Fig.\ref{fig: supple_qualitative_results_crowd_counting_2}.
With simulated training images and corresponding point annotations, we apply our SAE to restore the simulated point annotations. 
As shown in the right part of Fig.~\ref{fig: suppple_distribution}, the point annotations restored by SAE are closer to the distribution center, thereby exhibiting improved consistency compared with the original simulated point annotations.
To quantify the consistency of the point annotations, we compute the 2D position variances for the original simulated point annotations and the point annotations restored by SAE, respectively.
Through the SAE restoration, the position variance of point annotations is reduced from $0.223 \times r^2$ to $0.016 \times r^2$, 
quantitatively demonstrating the effectiveness of SAE in enhancing the consistency of point annotations.

\section{Pretraining of SAE}
\label{supple_sec: pretraining}
The Masked Autoencoders (MAE)~\cite{he2022mae} is trained by masking a portion of an image and reconstructing the masked image to the original unmasked one. This trained MAE backbone can be utilized to enhance the models of downstream tasks. 
Similarly, the proposed SAE is trained by restoring the shifted point annotations to their original positions. 
This process could be considered as a form of representation learning for object counting like MAE~\cite{he2022mae}.
Therefore, the trained SAE also provides a pretrained backbone for counting models.
We validate its effectiveness on two counting methods (one density-map-based and one localization-map-based) that employ the same VGG16~\cite{simonyan2014very_vgg} backbone with the SAE network.

As shown in Tab.~\ref{table: pretraining}, utilizing the pretrained backbone of the SAE alone consistently enhances the counting performance of CSRNet~\cite{li2018csrnet} and P2PNet~\cite{song2021rethinking_P2P} on the SH PartA dataset~\cite{zhang2016single_SH_MCNN}.
While utilizing both the annotations restored by SAE and the pretrained backbone of SAE, further improvements in counting accuracies are achieved. 
Notably, P2PNet using both the SAE-revised annotations and the SAE pretrained backbone, decreases the Mean Absolute Error (MAE) and Mean Squared Error (MSE) from  52.8 and 85.8 to 47.2 ($\downarrow$ 5.6) and 75.7 ($\downarrow$ 10.1), respectively. 
This study demonstrates the effectiveness of the SAE pretrained backbone and the versatility of the SAE methodology for object counting.

\begin{table}[h]
\centering
  \caption{Experimental results of using the pretrained backbone of SAE and/or point restoration by SAE. Restoration: train with the point annotations restored by SAE; Pretraining: adopt the pretrained backbone of SAE.}
  \label{Table_pretraining}
  
  \resizebox{0.7\linewidth}{!}{
  \begin{tabular}{l||c c|c c}
    \specialrule{0.1em}{0pt}{0pt}
    \multirow{2}{*}{Method} & \multicolumn{2}{c|}{CSRNet~\cite{li2018csrnet}} &  \multicolumn{2}{c}{P2PNet~\cite{song2021rethinking_P2P}}  \\
    \cline{2-5}
			             &  MAE$\downarrow$ & MSE$\downarrow$ &  MAE$\downarrow$ & MSE$\downarrow$ \\
    
    \specialrule{0.1em}{0pt}{0pt}
    Baseline      &   64.0    & 110.3     &   52.8    & 85.8       \\
    \specialrule{0.1em}{0pt}{0pt}
\& Restoration            &   57.2 (\SAEcolor{$\downarrow$6.8})    & 96.8 (\SAEcolor{$\downarrow$13.5})   &   48.2 (\SAEcolor{$\downarrow$4.6})     &76.1 (\SAEcolor{$\downarrow$9.7})       \\
                        \& Pretraining             &   59.5 (\SAEcolor{$\downarrow$4.5})    & 100.4 (\SAEcolor{$\downarrow$9.9})     & 49.7 (\SAEcolor{$\downarrow$3.1})    & 81.6 (\SAEcolor{$\downarrow$4.2})     \\
                     \& Restoration \& Pretraining          &   56.9 (\SAEcolor{$\downarrow$7.1})    & 92.1 (\SAEcolor{$\downarrow$18.2})      &   47.2 (\SAEcolor{$\downarrow$5.6})    & 75.7 (\SAEcolor{$\downarrow$10.1})   \\
  \specialrule{0.1em}{0pt}{0pt}
\end{tabular}}
\label{table: pretraining}
\end{table}

\section{Visualization}
\label{supple_sec: visualization}
\begin{figure}[t]
\centering

\subfloat{\includegraphics[width=.48\linewidth]{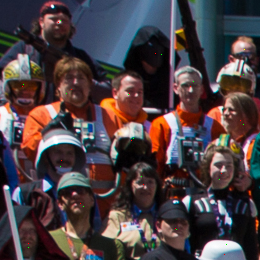}}\hspace{-2pt}
\subfloat{\includegraphics[width=.48\linewidth]{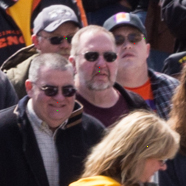}}

\subfloat{\includegraphics[width=.48\linewidth]{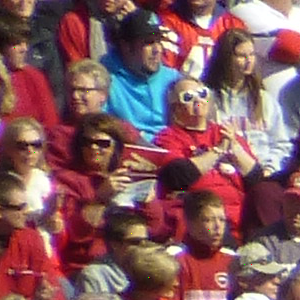}}\hspace{-2pt}
\subfloat{\includegraphics[width=.48\linewidth]{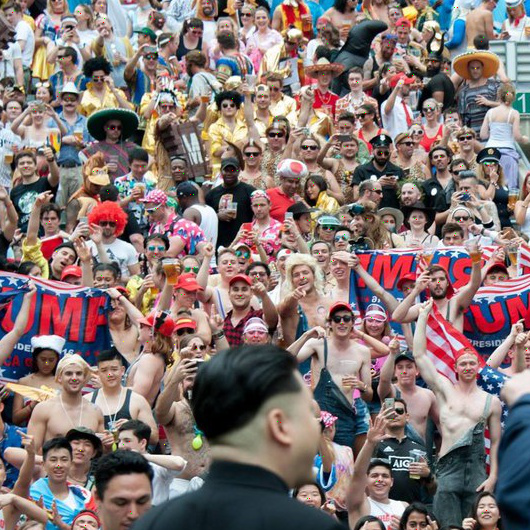}}

\subfloat{\includegraphics[width=.48\linewidth]{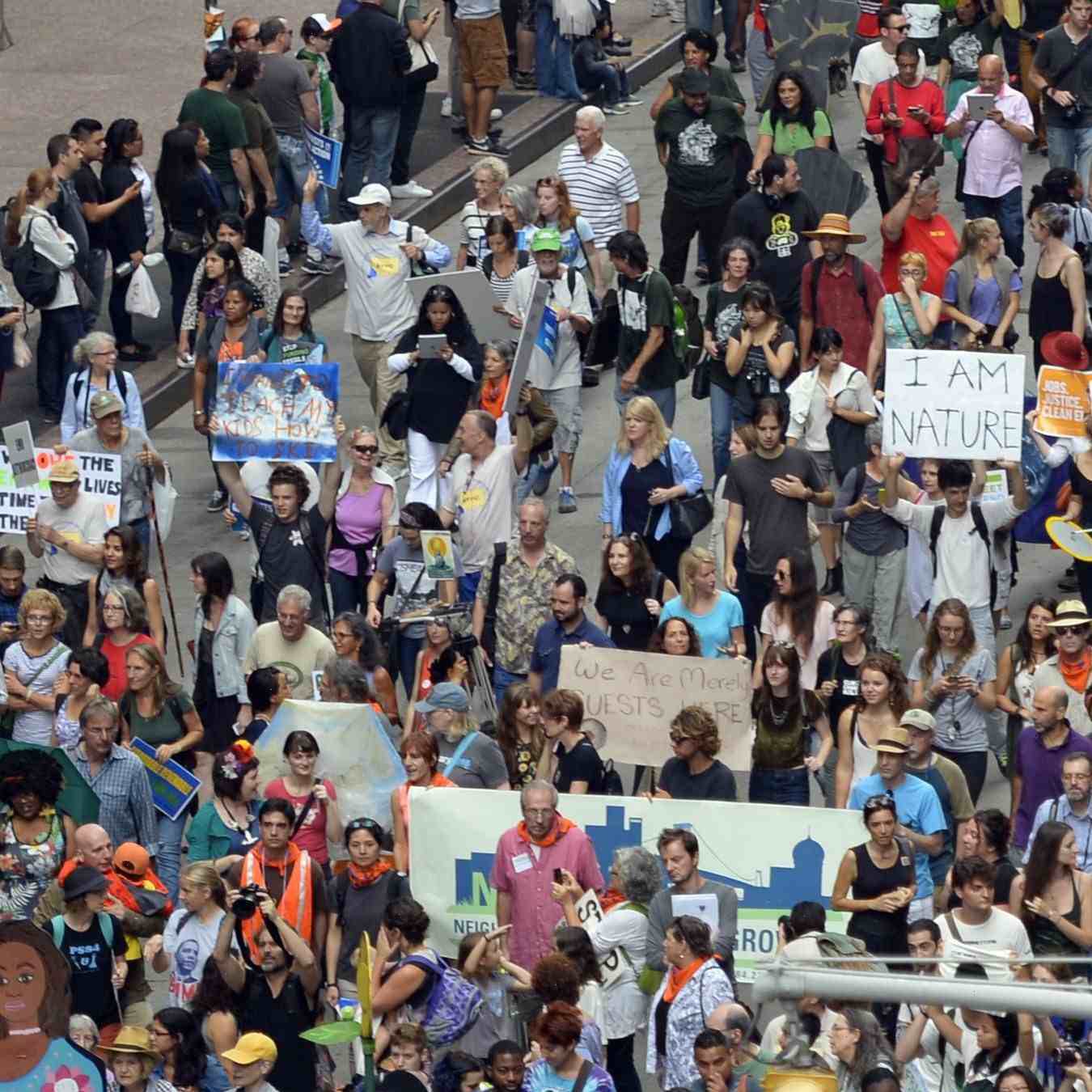}}\hspace{-2pt}
\subfloat{\includegraphics[width=.48\linewidth]{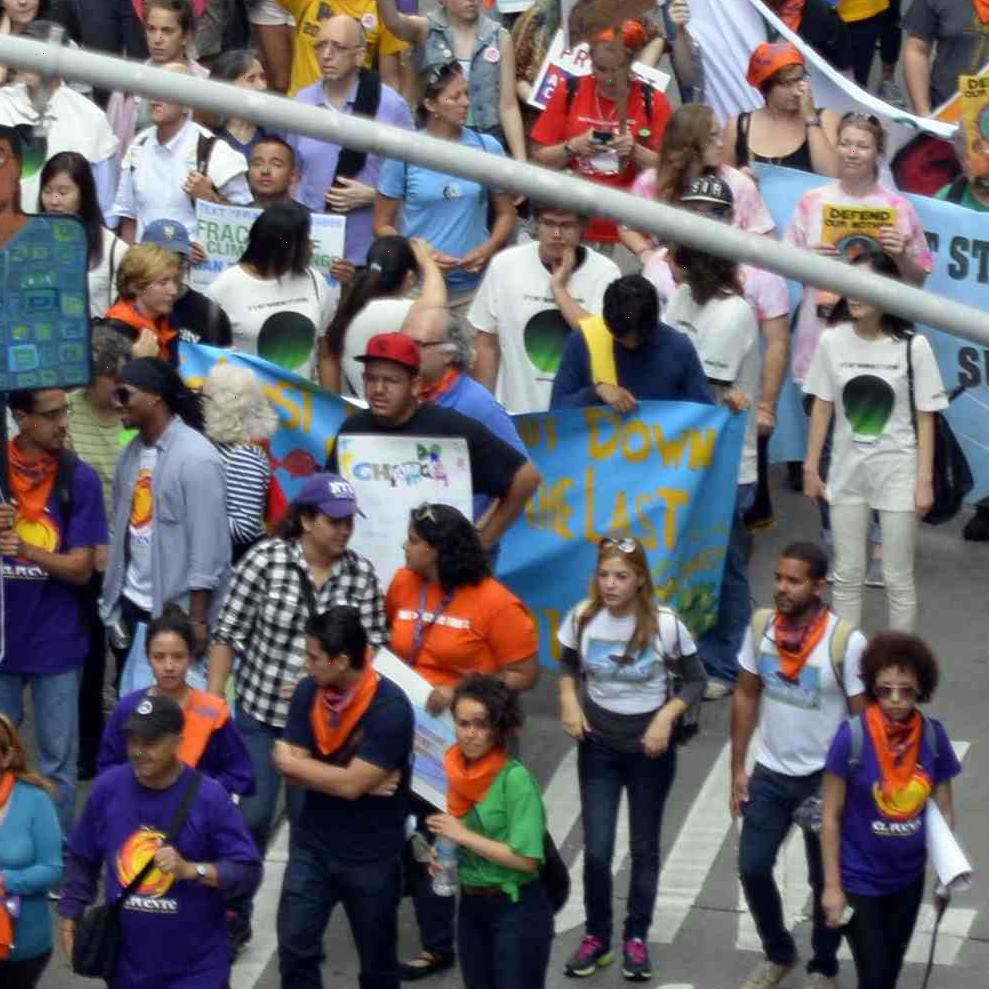}}

\caption{Visualization of restored point annotations provided by the SAE for crowd counting datasets. Zooming in is recommended for better readability in digital format.
}
\label{fig: supple_qualitative_results_crowd_counting_1}
\end{figure}

\begin{figure}[t]
\centering

\subfloat{\includegraphics[width=.47\linewidth]{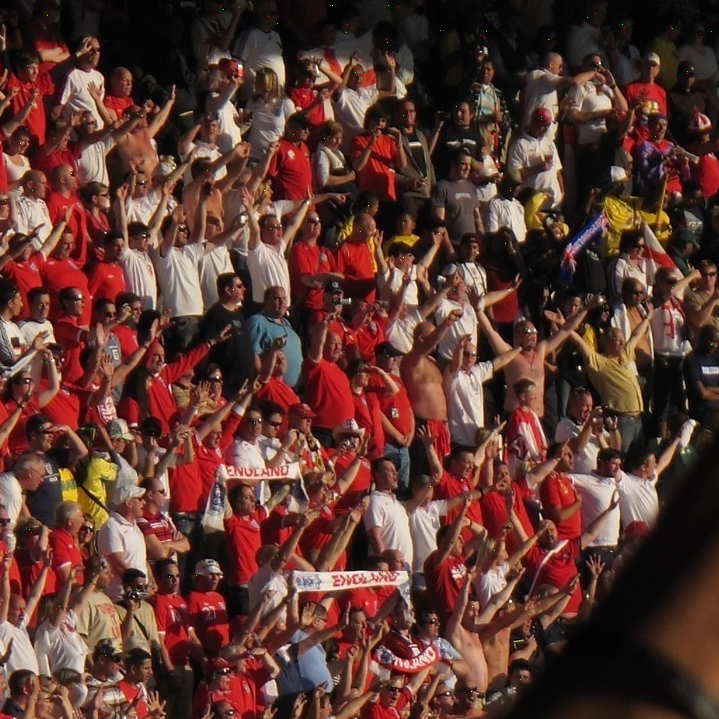}}\hspace{-2pt}
\subfloat{\includegraphics[width=.47\linewidth]{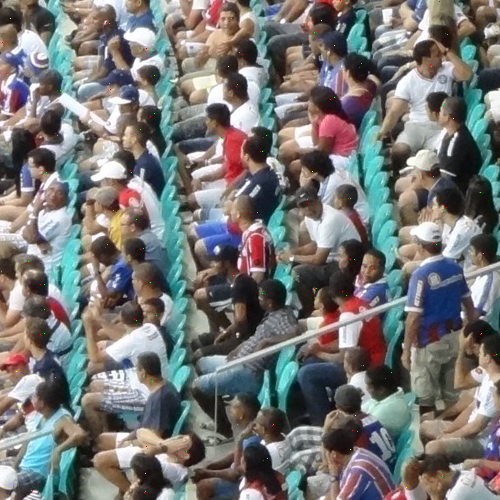}}

\subfloat{\includegraphics[width=.47\linewidth]{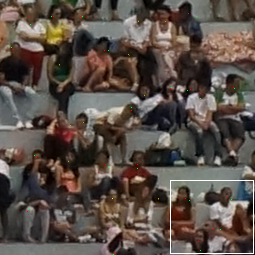}}\hspace{-2pt}
\subfloat{\includegraphics[width=.47\linewidth]{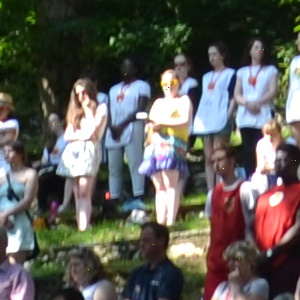}}

\subfloat{\includegraphics[width=.47\linewidth]{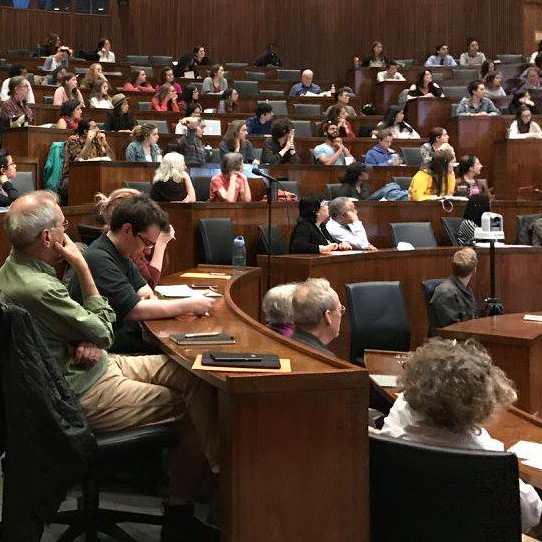}}\hspace{-2pt}
\subfloat{\includegraphics[width=.47\linewidth]{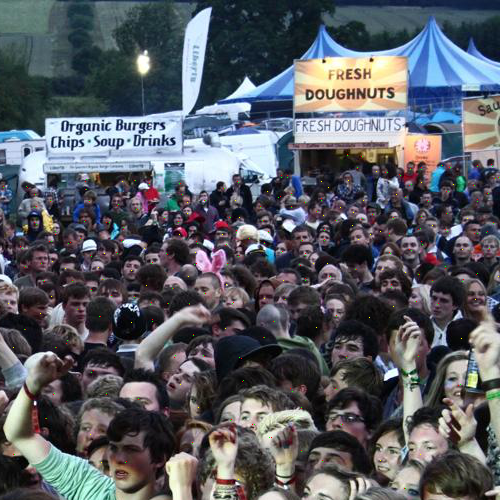}}
\caption{Visualization of restored point annotations provided by the SAE for crowd counting datasets. The white boxed area in the middle left image presents some heads with point annotations outside the head regions. Zooming in is recommended for better readability in digital format.
}
\label{fig: supple_qualitative_results_crowd_counting_2}
\end{figure}

\begin{figure}[t]
\centering

\subfloat{\includegraphics[width=.48\linewidth]{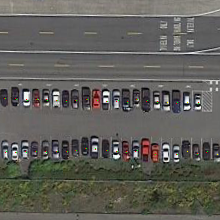}}\hspace{-2pt}
\subfloat{\includegraphics[width=.48\linewidth]{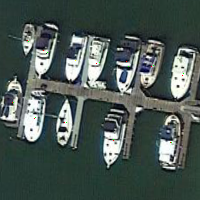}}

\subfloat{\includegraphics[width=.48\linewidth]{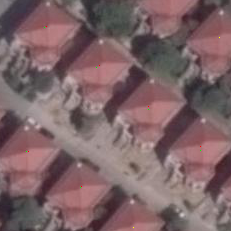}}\hspace{-2pt}
\subfloat{\includegraphics[width=.48\linewidth]{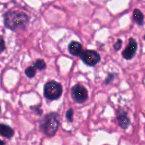}}

\subfloat{\includegraphics[width=.48\linewidth]{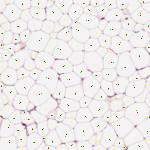}}\hspace{-2pt}
\subfloat{\includegraphics[width=.48\linewidth]{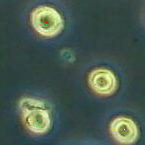}}

\caption{Visualization of restored point annotations provided by the SAE for remote sensing object counting datasets and cell counting datasets. Zooming in is recommended for better readability in digital format. }
\label{fig: supple_qualitative_results_remote_sensing_object_and_cell_counting}
\end{figure}

We illustrate some more visualization results of restored point annotations given by the proposed SAE in Fig.~\ref{fig: supple_qualitative_results_crowd_counting_1} \ref{fig: supple_qualitative_results_crowd_counting_2} \ref{fig: supple_qualitative_results_remote_sensing_object_and_cell_counting}. 
\textcolor{green}{Green} points: initial point annotations; \textcolor{red}{Red} points: revised point annotations; \textcolor{yellow}{Yellow} points: initial point annotation coincides with the corresponding revised point annotation.

